%% file: templateArxiv.tex
\documentclass{article}

\usepackage{PRIMEarxiv}

\usepackage[utf8]{inputenc} % allow utf-8 input
\usepackage[T1]{fontenc}    % use 8-bit T1 fonts
\usepackage{hyperref}       % hyperlinks
\usepackage{url}            % simple URL typesetting
\usepackage{booktabs}       % professional-quality tables
\usepackage{amsfonts}       % blackboard math symbols
\usepackage{nicefrac}       % compact symbols for 1/2, etc.
\usepackage{microtype}      % microtypography
\usepackage{lipsum}
\usepackage{fancyhdr}       % header
\usepackage{graphicx}       % graphics
\usepackage{subfigure}
\usepackage{booktabs} % for professional tables
\usepackage{physics}
\usepackage{algorithm}
\usepackage{algpseudocode}
\usepackage{mathtools}
\usepackage[monochrome]{xcolor}
\usepackage{multirow}

\usepackage{hyperref}

\usepackage{amsmath}
\usepackage{amssymb}
\usepackage{mathtools}
\usepackage{amsthm}
\usepackage{empheq}
\usepackage{breqn}

\usepackage[capitalize,noabbrev]{cleveref}

\theoremstyle{plain}
\newtheorem{theorem}{Theorem}[section]

\theoremstyle{definition}

\theoremstyle{remark}

\allowdisplaybreaks

\title{Hidden Traveling Waves bind Working Memory Variables
in Recurrent Neural Networks
}

\author{
  Arjun Karuvally, \\
  College of Information and Computer Sciences \\
  University of Massachusetts \\
  Amherst MA, USA\\
  \texttt{akaruvally@umass.edu} \\
   \And
  Terrence J. Sejnowski \\
  Computational Neurobiology Laboratory, \\
The Salk Institute for Biological Studies. \\
  La Jolla CA, USA \\
  \texttt{terry@snl.salk.edu} \\
  \And
  Hava T. Siegelmann, \\
  College of Information and Computer Sciences \\
  University of Massachusetts \\
  Amherst MA, USA\\
  \texttt{hava@umass.edu} \\
}

\begin{document}
\maketitle

\begin{abstract}
Traveling waves are a fundamental phenomenon in the brain, playing a crucial role in short-term information storage. In this study, we leverage the concept of traveling wave dynamics within a neural lattice to formulate a theoretical model of neural working memory, study its properties, and its real world implications in AI. The proposed model diverges from traditional approaches, which assume information storage in static, register-like locations updated by interference. Instead, the model stores data as waves that is updated by the wave's boundary conditions. We rigorously examine the model's capabilities in representing and learning state histories, which are vital for learning history-dependent dynamical systems. The findings reveal that the model reliably stores external information and enhances the learning process by addressing the diminishing gradient problem. To understand the model's real-world applicability, we explore two cases: linear boundary condition and non-linear, self-attention-driven boundary condition. The experiments reveal that the linear scenario is effectively \textit{learned} by Recurrent Neural Networks (RNNs) through backpropagation when modeling history-dependent dynamical systems. Conversely, the non-linear scenario parallels the autoregressive loop of an attention-only transformer. Collectively, our findings suggest the broader relevance of traveling waves in AI and its potential in advancing neural network architectures.
\end{abstract}

% keywords can be removed

\input{main_paper}

%Bibliography
\bibliographystyle{unsrt}  
\bibliography{references}  

\newpage

\appendix

\input{appendix}

\end{document}

%% file: main_paper.tex
\section{Introduction}

Traveling waves are ubiquitous in neurobiological experiments of human memory \cite{Davis2020SpontaneousTW}.
They have been observed during awake and sleep states throughout the brain, including the cortex and hippocampus and have been shown to impact behavior \cite{Davis2020SpontaneousTC}.
Several hypotheses based on experimental evidence point to the utility of these waves in memory storage \cite{Benigno2023WavesTO}.
One of the hypothesis suggests external stimuli interacts with neural networks to form neural activity that propagates through the brain \cite{Muller2018CorticalTW,Perrard2016WaveBasedTM}.
A snapshot of this wave field in the brain provide the information necessary to reconstruct the recent past, a mechanism ideal for the memory storage.
Recent empirical evidence also points to the potential utility of traveling waves in artificial intelligence. 
Of note is a recent empirical work that introduced traveling waves explicitly in the parameters of Recurrent Neural Networks (RNNs) and found that this simple addition improves its learnability and generalization properties \cite{Keller2023TravelingWE}.

\begin{figure}[t]
    \centering
    \includegraphics[scale=0.65]{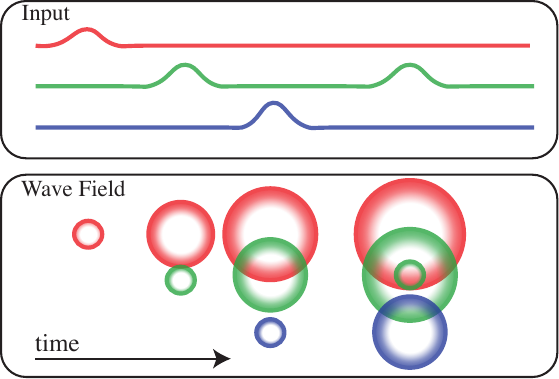}
    \caption{\textbf{Illustration of information storage in traveling waves} - A prominent hypothesis on the computational utility of traveling waves says that information is stored as ripple like waves that propagate outwards with time. A snapshot of the resulting wavefield provides all information necessary to reconstruct the recent history \textcolor{red}{by encoding both \textit{when} and \textit{where} (in the dimensions of the stimulus) a stimulus occured.}}
    \label{fig:bioTravelingWaves}
\end{figure}

Current hypotheses of working memory in RNNs, however, assume variables as bound in fixed register-like locations \cite{Sreenivasan2014RevisitingTR,Chumbley2008AttractorMO,Compte2000SynapticMA} that are updated by interference \cite{McGeoch1932ForgettingAT} or decay over time \cite{Barrouillet2011FurtherEF} to make way for new variables.
In this work, we assume the alternate principle that working memory variables are bound to traveling waves of neural activity.
We first derive a working memory model using the traveling wave principle applied to a neural substrate. We show that this model can approximate the state history required to encode any history-dependent dynamical system. We demonstrate the practical utility of the model by showing theoretical connections between the model and RNNs by assuming linear wave boundary conditions, and the autoregressive loop of transformer self-attention by assuming non-linear wave boundary condition. We then show theoretical and empirical evidence for the existence of traveling waves in trained Recurrent Neural Networks. Further, we also demonstrate the benefits of utilizing traveling waves in the gradient propagation behavior of RNNs.
The results demonstrate the applicabiltiy of traveling waves beyond neurobiology, and its potential in understanding and improving the computational properties of Recurrent Neural Networks.

\section{History-dependent Dynamical Systems (HDS)}

\textcolor{red}{Working memory in humans is typically studied using list recall tasks \cite{Wechsler1945ASM,Cabbage2017AssessingWM}}. In a typical setting for these tasks, list items to be remembered are first presented to a subject, then a math distractor is presented to remove any primacy or recency effects, and finally the subject is asked to recall the items in the presented list.
This study protocol provides sufficient and useful information about human working memory, which we know very less about, but investigating AI memory systems require tasks that are more challenging and closer to what the AI is used for. \textcolor{red}{Consider the example of a question answering neural network typically used in Natural Language Processing applications. At discrete timesteps, the user first provides the question and then the answer is obtained by running the network. This setting widely differs from that of the list recall task.}

To address the issue of practicality, while retaining experimental control, we consider a class of discrete history-dependent dynamical systems (HDS) as tasks to experimentally and theoretically evaluate working memory. HDS are a class of dynamical systems where the the next state depends on the current state \textit{and} the state history prior to the current state.
A canonical example of an HDS is the Fibonacci series, where the next state is the arithmetic sum of the previous two states of the system.
The HDS is initialized with the necessary history to start generating future states without any ambiguity akin to the query that is provided to a generative model.
Compared to the traditional working memory tasks, HDS have the following beneficial properties as experimental and theoretical setups - (1) their dynamic equations can be known in advance enabling reasoning about any learned behaviors, (2) their parameters such as history and evolution function can be experimentally controlled, (3) they have infinite horizon enabling evaluating length generalization properties.

Mathematically, the state evolution of a general history-dependent dynamical system with a history of $s$ states is represented by the following equations.
\begin{equation}
    \begin{cases}
        x_i & \text{if} \, 1 \leq i \leq s \\
        x_{i} = f(x_{i-1}, \hdots, x_{i-s}) &  \text{if} \, i > s
    \end{cases}
    \label{eqn:HDSDefinition}
\end{equation}
The first $s$ states of the system are initialized with $x_i \in \mathbb{R}^d$, where $d$ is the dimension of the state space. After the $s$ steps, the system evolves according to the rule specified by the function $f$ acting on the previous $s$ states. For example, in the canonical Fibonacci case, $s=2$ and the function $f$ is the addition operation. Now that we have a controlled experimental setup for testing working memory, we define the model in the next section.

\section{Traveling Wave based Working Memory (TWM)}

\begin{figure}
    \centering
    \includegraphics[scale=0.65]{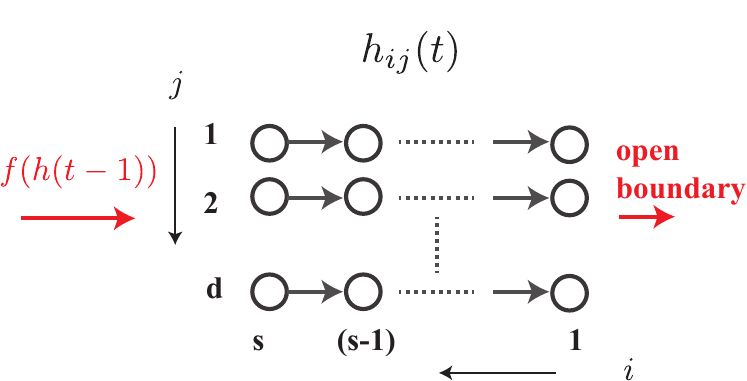}
    \caption{\textbf{Traveling Wave Memory Architecture} - The traveling wave based working memory architecture is composed of a neural substrate $h_{i j}$ with neurons arranged in a rectangular lattice. $d$ independent waves travel from the column with index $i=s$ down to $i=1$. These waves are independent and do not interact with each other as they travel in the substrate. The end boundary of the substrate is left open so there is no interference from reflecting waves, and the start boundary condition is computing a function $f$ of the entire neural substrate at the previous time step. This simple model is found to underlie working memory storage in RNNs.}
    \label{fig:twmArchitecture}
\end{figure}

Modeling of an HDS requires efficient storage of the past $s$ states in a manner that is accessible to compute future states. 
Further, the storage mechanism should enable updating these past states as needed.
We use the traveling wave principle to formulate a working memory model where each dimension of the state history is represented by 1-D traveling wave of activity in a neural substrate as shown in Figure \ref{fig:twmArchitecture}. 

Two components are required to fully define any wave propagation behavior in a substrate - (1) Wave components and how they interact as the waves travel in the substrate (2) Conditions for what happens to the waves at the boundaries of the substrate.
For the first component, we consider each dimension of the state as an independent wave that does not interact with the other waves as it travels in the substrate.
The interactions are at the the start boundary computing the function $f$ and generating new states to propagate. The end boundary is left open so that the waves do not reflect and interfere.

To derive the working memory model with the two components, we start with the equations for waves traveling in a continuous substrate and discretize the substrate and time.
1-D waves traveling in $d$ independent continuous substrates (represented by the position $v_j, j \in \{ 1, 2, \hdots, d \}$) is given by the following equation.
\begin{equation}
    \pdv{h(v_j, t)}{t} = \nu_{j} \pdv{h(v_j, t)}{v_j}
\end{equation}
There is one wave for each dimension of the state. Now, we Euler discretize $v_{j}$ with a step size of $1$ to account for the discrete positions of neurons in the wave substrate, Euler discretize time with step size of $1$ to obtain a discrete dynamical system, and apply the function $f$ at the start boundary with the end boundary open. We finally arrive at the following discrete dynamics (Appendix \ref{Appendix:TWMDefinition}).
\begin{equation}
    \begin{cases}
        h_{i, j}(t+1) = h_{i-1, j}(t) & 1 \leq i < s \\
        h_{s, j}(t+1) = f(h_{s}(t), \hdots, h_{1}(t))_{j} & \text{otherwise}
    \end{cases}
    \label{eqn:discreteTWM}
\end{equation}
The wave activity begins at the start location $s$ as a vector of neural activities and travels through the substrate till column $1$ where the open boundary makes the activity flow out without reflection.
As the wave flows, new activity is added to location $s$ as a function $f$ of the entire wave state which gets propagated. 
This process can potentially have infinite horizon. 
%
%In typical applications, we are not interested in only the activity of a subset of neurons. Thus, we can add a linear transformation that "reads" only the beginning of the wave. 
%
%The final traveling wave working memory has the following dynamical equations.
%

\begin{theorem}
Any history dependent dynamical system with a state dimension of $d$, a history of $s$ states and an evolution function $f$ can be represented in the traveling wave model.
\end{theorem}
The proof is by construction. For all HDS, the state history is defined as the neural activity $h$ and the function $f$ computing the start boundary of the traveling wave is the same as the function $f$ computing the next state in the HDS.

To illustrate this equivalence, consider an elementary example of encoding the Generalized Fibonacci series where each element $F_n \in \mathbb{R}^d$ is a vector defined recursively as,
\begin{equation}
    F_n = \begin{cases}
        u^1 & n=1 \\
        u^2 & n=2 \\
        \vdots & \\
        u^s & n=s \\
        \sum_{t=n-s}^{n-1} F_t & n > s \,
    \end{cases}
    \label{eqn:GeneralizedFibonacci}
\end{equation}
for any $u^i \in \mathbb{R}^d$. In order to store this \textit{process} of generating sequences of $F_n$, the vectors $u_1, u_2, \hdots u_s$ needs to be stored as variables and recursively added to produce new outputs. 
In the wave working memory framework, this can be accomplished by initializing the neural substrate such that $h_{i j} = (u^i)_j$, that is each $u$ is stored as activity of distinct columns of the substrate.
To encode the Fibonacci process in $\Phi$, the end boundary is open and the start boundary is defined by the following equation.
\begin{equation}
    f(h_{s-1}(t), \hdots, h_{1}(t))_{j} = \sum_{i=1}^s h_{i, j}
\end{equation}
The construction theorem and the Fibonacci example demonstrates that the general TWM stores and reliably computes the past information for any HDS. We now consider two cases of the TWM - linear boundary condition (LBC) and self-attention boundary condition (SBC) that are relevant for practical AI applications.

\subsection{Linear Boundary Condition (LBC)}

The Generalized Fibonacci was an elementary example whose purpose was to demonstrate how a given HDS can be converted to the traveling wave form. We now make two key generalizations of the Generalized Fibonacci setup that makes the traveling wave model relevant to practical neural models. The \textit{first generalization} focuses on the nature of the function $f$. In Generalized Fibonacci, $f$ was a simple addition operation over all the previous stored history states. In the LBC, the $f$ is generalized to be any \textit{linear} operator acting on the previous history states. The \textit{second generalization} is related to the linear basis for representing the model.
In the Generalized Fibonacci example, we implicitly assumed the standard basis for the substrate state representation. We allow for representing the LBC in any linearly independent basis vectors.
We show that with these two generalizations, the traveling wave model has a simple iterative matrix representation applicable to RNN dynamics.

LBC is defined with a flattened state representation (column wise) for the neural substrate $h'_{id+j} = h_{i, j} \psi_{id+j}$, where the set of column vectors $\psi_{id+j} \in \mathbb{R}^{sd}$ form a basis. With the linear $f$ assumption and the new basis, we can represent the entire traveling wave dynamics as a simple iterative application of a matrix shown below. 
\begin{equation}
    \Phi = \sum_{\mu=1}^{(s-1) d} \psi_{\mu} \psi^{\mu+d} + \underbrace{\sum_{\mu=(s-1)d+1}^{s d} \sum_{\nu=1}^{s d} \Phi_{\nu}^{\mu} \psi_\mu \psi^\nu}_{f(u(t-1), u(t-2), \hdots, u(t-s))} \, .
    \label{eqn:PhiTheoretical}
\end{equation}
The matrix is represented as the sum of products of column and row vectors to conveniently separate out the two distinct operations of the matrix. \textcolor{red}{The first part encodes the wave propagation in a neural substrate defined in a space spanned by the basis elements $\psi_{\mu}$. The action of the first part of the matrix in this space is visualized as a shift operation in the first $s-1$ subspaces of Figure \ref{fig:VMInterpretability} (rows indexed $1-7$). Each basis element $\psi_{\mu}$ has associated with it a row vector, $\psi^{\mu}$ (note the change from subscript to superscript) defined such that the dot product of the row with the column basis vector, $\langle \psi^{\mu},  \psi_{\nu} \rangle = 1$ only if $\mu=\nu$, else $\psi^{\mu} \psi_{\nu} = 0$.
The second part encodes the linear boundary condition $f$ (the $8^{\text{th}}$ row) of Figure \ref{fig:VMInterpretability}. This matrix acts on the entire \textit{flattened} wave substrate and store the result in the $s^{\text{th}}$ column of the substrate.
The traveling wave model now can be represented as a simple iterative application of the matrix $\Phi$ on the flattened substrate state $h$ : $h(t) = \Phi \, h(t-1)$.
Since the 'output' of the HDS encoded by the traveling wave model is only the $s^{\text{th}}$ column of the neural substrate, we add a transformation of the hidden state for output ($y(t)$)}. The other columns store the necessary history to compute the next state in the HDS, but is otherwise not relevant as output. The final dynamical evolution equations are shown below:
\begin{equation}
    \begin{dcases}
        h(t) = \Phi \, h(t-1) \, , \\
        y(t) = \sum_{(s-1)d+1}^{sd} e_{\mu - (s-1)d} \, \psi^{\mu} \, h(t) \, .
    \end{dcases}
    \label{eqn:LinearizedTWM}
\end{equation}
The final equations of the system has a familiar form to a first order approximation of RNN dynamics (Appendix \ref{appendix:TMVB:generalRNNs})!
In our experiments, we show that this similarity to RNN is not an artifact of the linear assumptions of the theory, but linearizing RNNs trained on HDS using backpropagation consistently converge to the TWM with LBC.
Another interesting connection of this wave dynamical equations is with existing state space models \cite{Gu2022HowTT, Gu2023MambaLS} where $\Phi$ is the matrix $A$. A prominent state space model, H3 \cite{Dao2022HungryHH}, has its $A$ matrix initialized with an off diagonal shift matrix - a variant of the wave propagation matrix with both the two boundaries open.

\subsection{Self-attention Boundary Condition (SBC)}

The linear boundary condition significantly restricts the type of HDS the traveling wave working memory can handle to linear cases. In this section, we consider the non-linear case, where the function $f$ is the self attention operator.
Consider the discrete TWM defined in Equation \ref{eqn:discreteTWM} with a self attention based dynamical evolution equations defined below
The boundary condition of this system can be written as:
\begin{equation}
		h_{s, j}(t+1) = \sum_{i=1}^{d} \frac{\exp( h_{i}^\top W_K W_Q h_{s})}{\sum_{k} \exp( h_{k}^\top W_K W_Q h_{s})} W_{V} h_{i} \\
\end{equation}
%
%If the
The resultant dynamical equations is the autoregressive computations of a single head of a attention layer in transformers.
This new perspective provides a justification for the transformer architecture, where the autoregressive loop of a self attention head is implicitly a traveling wave based working memory with non-linear boundary conditions.

This perspective provides alternate justification for the better properties (and applicability) of transformers compared to typical RNNs. One benefit is that, while RNNs have to \textit{learn} the traveling wave mechanisms through backpropagation, transformers are set up to explicitly encode the history in its context. This significantly reduces the training time and prevents potential issues arising from poor traveling wave encoding in RNNs.
Another tentative benefit comes in the form of improved non-linear interactions and the ability to approximate any non-linear boundary conditions using the self attention operation with learnable matrices $W_K, W_Q$ and $W_V$.
Taken together, traveling waves have far reaching implications beyond 
the theoretical benefits we showed here and can aid in the development of efficient and smarter AI systems.

\begin{table*}[t]
\includegraphics[scale=0.85]{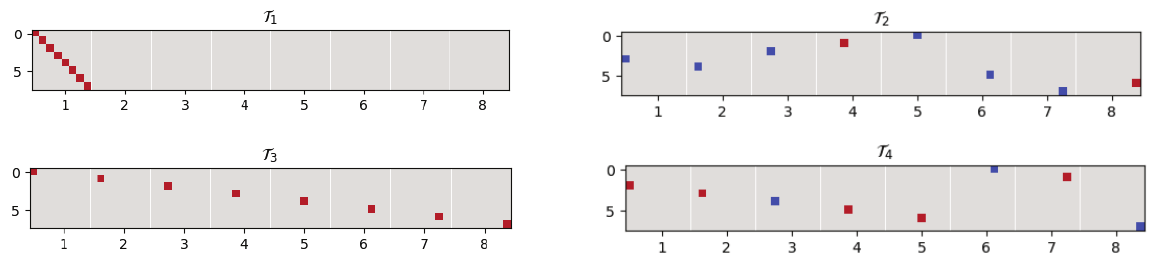}
\centering
\begin{tabular}{l|l|ll|ll}
\toprule 
    Model & Task & \multicolumn{2}{c}{hidden size: 64} & \multicolumn{2}{c}{hidden size: 128}\\
    &  & L2: 0.0 & L2: 0.001 & L2: 0.0 & L2: 0.001 \\
    \midrule
     \multirow{ 4}{*}{RNN (no bias)} & $\mathcal{T}_1$ & --- (0.97)& --- (0.88)& 0.0005  (1.00)& --- (0.93) \\
& $\mathcal{T}_2$ & 0.0075  (1.00)& --- (0.85)& 0.0055  (0.98)& 0.0031  (0.98)\\
& $\mathcal{T}_3$ & 0.0026  (1.00)& 0.0010  (0.97)& 0.0031  (0.98)& 0.0005  (1.00)\\
& $\mathcal{T}_4$ & 0.0112  (0.94)& 0.0011  (1.00)& 0.0022  (1.00)& 0.0006  (1.00)\\
\hline
   \multirow{ 4}{*}{RNN (with bias)} &$\mathcal{T}_1$     & --- &  --- &  0.0005  (1.00) &  --- \\ 
    &$\mathcal{T}_2$     & 0.0152  (1.00) &  --- &  0.0053  (1.00) &  0.0006  (1.00) \\ 
    &$\mathcal{T}_3$     & 0.0032  (1.00) &  0.0007  (1.00) &  0.0031  (1.00) &  0.0005  (1.00) \\ 
    &$\mathcal{T}_4$     & 0.0115  (1.00) &  0.0011  (1.00) &  0.0023  (1.00) &  0.0006  (1.00) \\ 
\bottomrule
\end{tabular}
\caption{\textbf{RNNs consistently converge to the TWM encoding of the HDS}: The top image shows the composition functions for the $4$ linear HDS tasks, visualized as a matrix with x-axis input, and y-axis output. Red color denotes +1, blue is -1 and no color is 0. $\mathcal{T}_1$ is the linear function representing the repeat copy task, the rest are other general variable composition functions. 
The matrix can be imagined as an operator acting on the $sd = 8 \times 8 = 64$ dimensional substrate vector from the left.
The table shows the MAE in the complex argument between the eigenspectrum of the predicted $\Phi$ from the TWM and the empirically learned $W_{hh}$ in the $4$ tasks across $20$ seeds under different RNN configurations. This average error is indeterminate (---) if the rank of the theoretical $\Phi$ is different from the empirical $W_{hh}$. Values in the brackets show the average test accuracy of the trained model. For models that have high test accuracy ($>0.94$), the error in the theoretically predicted spectrum is very low indicating consistent convergence to the theoretical circuit. A notable exception of this behavior is $\mathcal{T}_1$ with hidden size$=64$ and $L2=0$, where the restricted availability of dimensions forces the network to encode variables in bottleneck superposition resulting in a low-rank representation of the solution. It is notable that the low-rank matrix is also a TWM, where the state history representation is compressed.}
\label{table:VBConvergence}
\end{table*}

\section{Results}
In this section, we investigate the potential role of the TWM theory in informing and improving Recurrent Neural Networks (RNNs). First, we show empirical evidence of convergence to the theoretical equations of $\Phi$ in Equation \ref{eqn:PhiTheoretical}, demonstrating the practical relevance of the theory in improving our understanding of RNN computations. 
Utilizing our understanding of TWM, we show RNNs encoding past information as hidden traveling waves, revealed only after accounting for the basis transformation introduced in the LBC section.
We also show a simple method that linearly transforms and reveals the underlying TWM matrix of Equation \ref{eqn:PhiTheoretical} hidden in the trained weights of RNNs.
Lastly, we obtain the theoretical result that the TWM framework has benefits for training, by alleviating the diminishing/exploding gradient problem in RNNs.

\begin{figure*}[t]
	% \begin{subfigure}[b]
         \centering
		\includegraphics[scale=0.67]{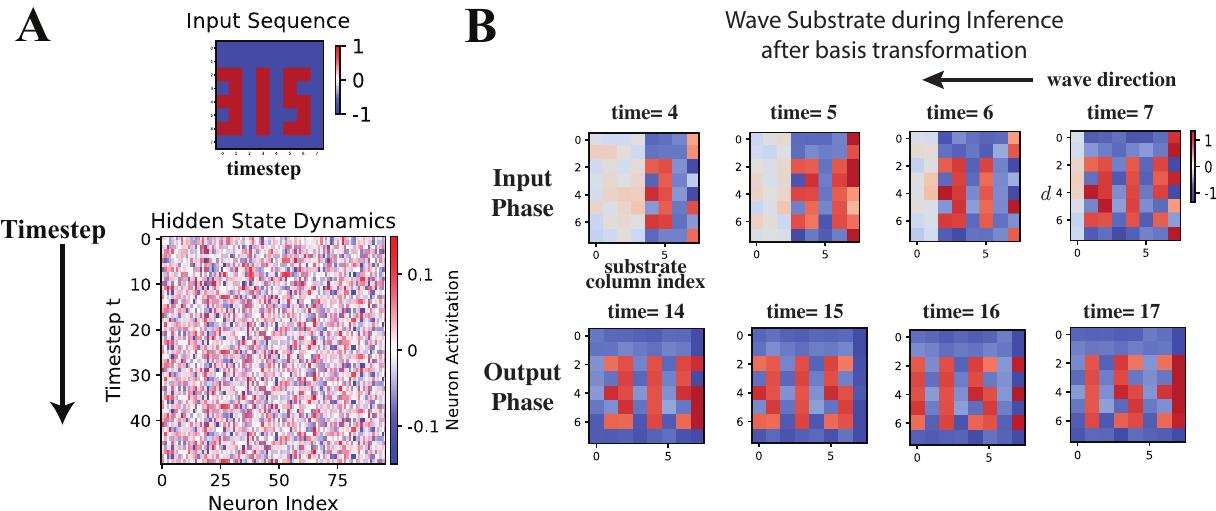}
     % \end{subfigure}
\caption{ \textbf{Basis transformation reveals traveling waves encoding the recent past in the repeat copy task (with $s=d=8$)}: \textbf{A}. In the repeat copy task ($\mathcal{T}_1$), the RNN needs to repeatedly produce an input sequence that is presented. A typical trained hidden state after providing the input does not show any meaningful patterns connected to the input. \textbf{B}. The same hidden states when their basis is transformed reveal the input information being stored as waves of activity traveling from the variable with index $8$ down to the variable with index $1$ that are repeatedly mutated with the boundary condition. }
\label{fig:VMHiddenNeurons}
\end{figure*}

\textbf{Experimental Setup}

For the experimental tasks, we consider linear restriction of HDS which ensures experimental control and ease of interpretation of learned recurrent behavior. The Generalized Fibonacci was a good theoretical setup, but cannot be learned directly in RNNs whose neural activity is typically bounded by squashing non-linearities. We therefore focus on a restricted HDS defined in Equation \ref{eqn:HDSDefinition} with binary state representations ($x_i \in \{ -1, 1 \}^d, f: \{ -1, 1 \}^{sd} \to \{ -1, 1 \}^d $). The function $f$ can be easily visualized as a matrix acting on the flattened neural substrate (Table \ref{table:VBConvergence}(top)). 
% Therefore, to compute the state at timestep $t>s$, a memory of $s$ previous states $x_{s}, \hdots, x_{1}$ needs to be stored.
%
To store the $s$ initial steps and to produce HDS without ambiguity, the entire learning setup is divided into two phases - the \textit{input phase} and the \textit{output phase}. 

The input phase lasts for $s$ timesteps. During each timestep $t$ (where $1 \leq t \leq s$) of this phase, the RNN receives a $d$-dimensional input column vector \( u(t) = [u^1(t), u^2(t), \hdots, u^d(t)]^\top \). These vectors \( u(t) \) provide the external information that the RNN is expected to store within its hidden state and initializes the HDS.

Once the input phase concludes at timestep $s$, the output phase begins immediately from timestep $s+1$. During the output phase, the RNN no longer receives external input and instead operates autonomously, generating outputs based on the information stored during the input phase. 

The RNN output is evaluated during the output phase and compared with the known ground truth of the HDS using the MSE loss. Backpropagation Through Time (BPTT) is employed as the learning algorithm to estimate the parameters of the RNN from the data. We do not restrict the RNN training in any other way. It is instructive to note that $\mathcal{T}_1$ (one of the HDS we consider) is the Repeat Copy task - a canonical task used to evaluate the memory storage properties of Recurrent Neural Networks. \textcolor{red}{See Appendix \ref{appendix:ERVB} for details on the experimental setup and methodology.}

\textbf{Model}

We consider simple Elman networks as RNN models for the experiments \cite{Rumelhart1986LearningRB}. Although elementary, these networks are canonical models to investigate the behavior of recurrent neural networks. Further, compared to advanced architectures like LSTMs, they provide ease of understanding and analytic simplicity. The Elmann RNN is defined by the following equations.
\begin{equation}
    \begin{cases}
        h(t) = \sigma(W_{h h} h(t-1) + W_{u h} u(t)) \\
        y(t) = W_r h(t)
    \end{cases}
\end{equation}
Here, $h(t) \in \mathbb{R}^n$ is a vector representing the hidden state of the RNN at time $t$, $u(t) \in \mathbb{R}^d$ is the input to the RNN at time $t$, and $y(t)$ is the output at time $t$. The $W_{h h} \in \mathbb{R}^{n \times n}, W_{u h} \in \mathbb{R}^{n \times d}, W_r \in \mathbb{R}^{d \times n}$ are matrices that are learned. The RNN non-linearity $\sigma$ is the tanh activation function.

\subsection{Linearization of RNNs trained on linear HDS show consistent convergence to Traveling Waves with LBC }
\textit{Do practical RNNs employ traveling waves to store history?} To answer this question, we trained various RNN configurations, differing in hidden sizes and regularization penalties on 4 HDS, each differing in the linear composition function $f$ (See matrix representation of $f$ at the top of Table \ref{table:VBConvergence}). After training, the RNNs were linearized, and the eigen-spectrum of the learned Jacobian matrix (Appendix \ref{appendix:TMVB:generalRNNs}) is compared with the theoretical $\Phi$, we found for the TWM with LBC in Equation \ref{eqn:PhiTheoretical}. If RNNs learn a representation in alignment with TWM, both operators, i.e., the learned $W_{hh}$ and theoretical $\Phi$, are expected to share a portion of their eigenspectrums as they are similar matrices (i.e they differ only by a basis change). \textcolor{red}{Note that it is possible to compare the theoretical $\Phi$ directly with the learned $W_{hh}$ due to the Jacobian of the $\tanh$ non-linearity being the identity matrix when linearized around the origin, which is a fixed point of the system. In general cases like the RNN with bias, the analysis has to account for the non-linearity by finding and linearizing around fixed points following the procedure in Appendix \ref{appendix:TMVB:generalRNNs}}. 

The comparison excludes the real part of the spectrum.
The rationale behind this exclusion lies in what the magnitude tells about the dynamical behavior. The eigenvalue magnitude portrays whether a linear dynamical system is diverging, converging, or maintaining consistency along the eigenvector directions \cite{Strogatz1994NonlinearDA}. RNNs typically incorporate a squashing non-linearity, such as the Tanh activation function, which restricts trajectories that diverge to infinity.
Essentially, provided the eigenvalue magnitude remains $\geq 1$, the complex argument solely determines the overall dynamical behavior of the RNN. 
Table \ref{table:VBConvergence} depicts the average absolute error in the eigenspectrum and test accuracy when the RNN models are trained across $4$ distinct variable binding tasks. The table shows that RNNs consistently converge to the hypothetical circuit. 
This indicates that RNNs employ traveling waves in their dynamics. 
Next, we investigate what information these traveling waves carry and how TWM with LBC can be utilized to analyze and understand RNN behavior.

\subsection{Hidden traveling waves encode recent history in trained RNNs}
We approximated the wave substrate of RNNs trained on the Repeat Copy task $(\mathcal{T}_1)$ using an algorithm and visualized the hidden state.
$\mathcal{T}_1$ was chosen because it is easy to interpret and visualize, the results hold for other binary HDS tasks also.
\textcolor{red}{We use a power iteration based algorithm described in Algorithm \ref{alg:variableMemories} to approximate the linear basis of the TWM with LBC learned by the empirically trained RNNs (detailed error analysis can be found in Appendix \ref{appendix:algorithmFailureModes})}.

\begin{figure*}[t]
	% \begin{subfigure}[b]
         \centering
		\includegraphics[scale=0.68]{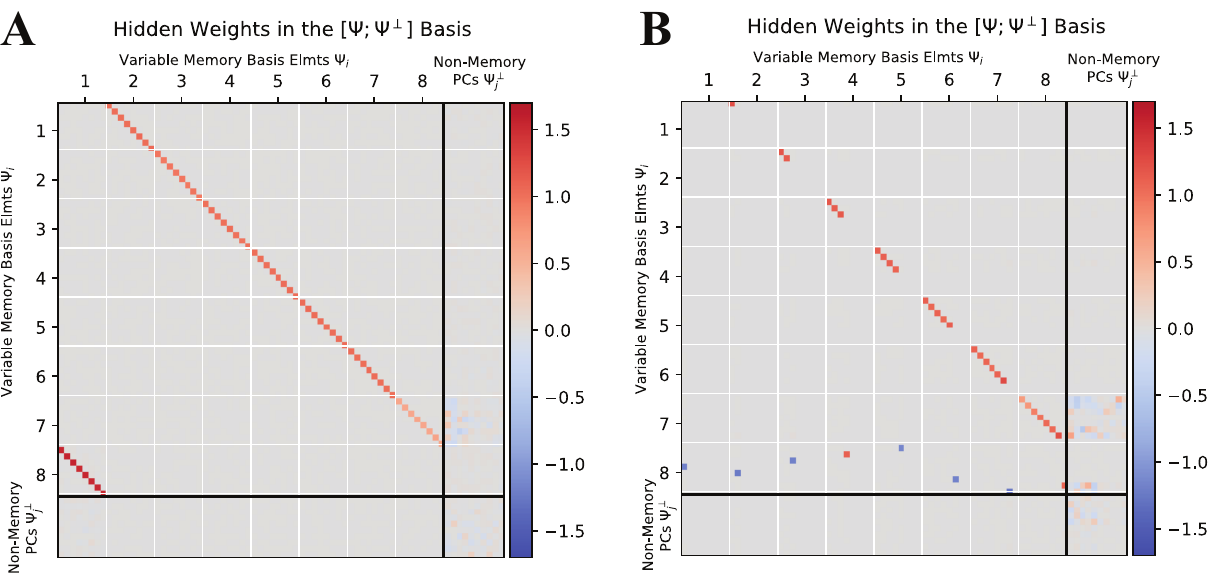}
     % \end{subfigure}
\caption{ \textbf{TWM improve human interpretation of the RNN parameters (with $s=d=8$)}: The learned weights when visualized in the LBC basis results in a form that is human-interpretable. For RNNs trained on two sample tasks $\mathcal{T}_1$ (\textbf{A} left) and $\mathcal{T}_2$ (\textbf{B} right), the weight matrix $W_{hh}$ converts into a form that reveals internal mechanisms of how RNNs solve the two tasks. For both tasks, the variables with index $<8$ copies its contents to the preceding variable resulting in a wave of activity. Variable $8$ actively computes the function $f$ applied on all the variables stored in the hidden state as boundary condition. For $\mathcal{T}_1$, the boundary condition is a simple copy of the $1^{\text{st}}$ variable, and for $\mathcal{T}_2$, it is a linear composition of all the variables Notably, the circuit for $\mathcal{T}_2$ shows an \textit{optimized basis} where the wave for each dimension travels only till the boundary that is necessary to be stored for computation. }
\label{fig:VMInterpretability}
\end{figure*}

\begin{algorithm}[t]
\caption{Algorithm to approximate linear basis of trained RNNs}\label{alg:variableMemories}
\begin{algorithmic}
\State $0 \leq \alpha \leq 1$
\State $s$ \Comment{number of time-steps in the input phase}
\State $W_{hh}, W_{r}$ \Comment{learned parameters of the RNN}
\State $\Psi_s \gets W_r^\dag$
\For{$k \in \{ s-1, s-2, \hdots 1 \}$}
     \State $\Psi_k \gets \left(\left( W_{hh}^\top \right)^k W_r^\dag \right)$
     \State $\Psi_k \gets \Psi_k - E E^\dag \Psi_k \quad \forall E: \lambda(E) < 1$ \Comment{Remove the components along transient directions.}
\EndFor
\State $\Psi \gets [\Psi_1;\hdots;\Psi_s]$
\State $\Psi^\perp \gets \text{PC}(\{\tilde{h(t)}\} - \Psi \Psi^\dag \, \{\tilde{h}(t)\})$ \Comment{Principle Components of $\tilde{h}$ from simulations}
\end{algorithmic}
\end{algorithm}

In the Repeat Copy task, the RNN must repeatedly output the stream of inputs provided during the input phase.
The simulated hidden states of learned RNNs are visualized by projecting the hidden state into the columns of the wave substrate: $\tilde{h} = \Psi \Psi^\dag h$, where $\dag$ is the Moore Penrose pseudo-inverse of the matrix $\Psi$.
The results shown in Figure \ref{fig:VMHiddenNeurons} reveal that the hidden state is in a superposition (or distributed representation) of latent neurons that actively store each variable required to compute the function $f$ at all points in time.
These variables act as the substrate for wave propagation as shown in the figure.

Finding the linear basis transformation also means that the hidden computations of the learned parameters can be revealed by using the transformation. 
Figure \ref{fig:VMInterpretability} utilizes the computed basis to transform the learned parameters of the RNN. The figure shows connectivity structure exactly predicted by the TWM with LBC encoding for each task.

\begin{figure*}[t]
	% \begin{subfigure}[b]
         \centering
		\includegraphics[scale=0.75]{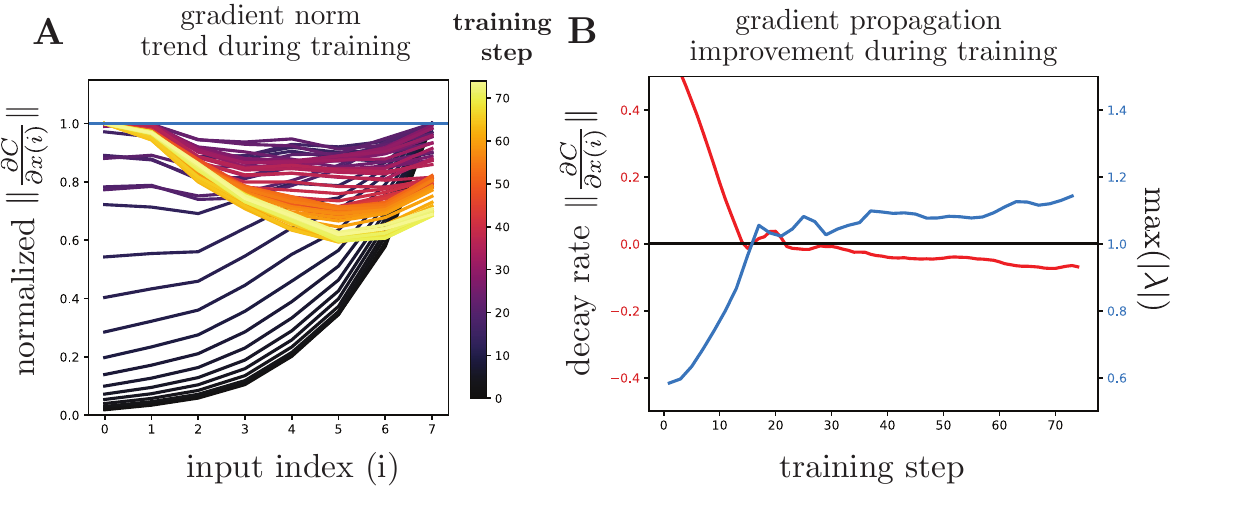}
     % \end{subfigure}
\caption{ \textcolor{red}{\textbf{Analysis of the RNN gradient propagation behavior reveals the diminishing gradient problem alleviating during training, as predicted by the TWM}: The experiment on the trends in the gradient norm with respect to RNN inputs show that the diminishing gradient issue reduces during training \textbf{A.} During early training iterations, the gradient norm decays exponentially the farther it is propagated (denoted by lower input indices). This diminishing behavior is slowly alleviated as training progresses, with later training iterations gradient norm approaching $1$ - the ideal gradient norm required for preventing the diminishing/exploding gradient issue. \textbf{B} The analysis of the decay rates (shown by the red line) during training show sharp decrease during training, after which the decay rate remains close to 0. The transition to the decay rate of 0 happens when the absolute value of the maximum eigenvalue (shown by the blue line) crosses 1. At this point, the eigenvalue crosses the unit circle in the imaginary plane and traveling waves are set up. Taken together, the two plots verify the predictions of the TWM in relation to the gradient propagation behavior of RNNs.} }
\label{fig:VMGradientPropagation}
\end{figure*}

\subsection{TWM improve gradient propagation}

A consistent problem in RNNs is the diminishing/vanishing gradient problem, which has been the subject of intense investigations and model improvements \cite{Hochreiter2001GradientFI,Pascanu2012OnTD}.
The main argument for the gradient problem is that the repeated application of the RNN weight matrix to the gradient results in the gradient magnitude diminishing or exploding the farther back it is propagated \cite{Arjovsky2015UnitaryER}.
TWMs provide an alternate solution to the problem by representing the state history spatially in the RNN hidden state without resorting to advanced architectures.
This representation thus removes the requirement for the gradient to be propagated backward in time, entirely eliminating the diminishing gradient problem.
Assume an RNN learning the history dependent dynamical system given by Equation \ref{eqn:HDSDefinition}.
The RNN is fed the initial conditions in the first $s$ timesteps and then, the output of the RNN at each timestep $y(t)$ is compared with the $x(t)$ in the equation, obtained by computing $f$ for $t>s$ timesteps.
The HDS is 

\textbf{Without TWM}

Let's consider a random initialization of the RNN devoid of any wave phenomenon.
Let $C(y(s+1), x(s+1))$ be the cost function computed at time $s+1$.
We will now analyze the influence of the input at time $1$ - $x(1)$ on the cost function evaluated at time $s+1$. 
\begin{align}
    \pdv{C}{x(1)} &= \pdv{C}{h(s+1)} \left( \prod_{k=1}^{s} \pdv{h(k+1)}{h(k)} \right) W_{xh} \\
     & = \pdv{C}{h(s+1)} \left( \prod_{k=1}^s \mathcal{J}(\sigma) W_{hh} \right) W_{xh} \\
\end{align}
Now, if we are to upper bound the norm of the cost gradient,
\begin{align}
    \lVert \pdv{C}{x(1)} \rVert &\leq \Vert \pdv{C}{h(s)} \Vert \, \Vert \left( \mathcal{J}(\sigma) \, W_{hh}\right) \Vert^s \, \Vert W_{xh} \Vert \\
    &\leq \Vert \pdv{C}{h(s)} \Vert \, \Vert W_{xh} \Vert \, \mu^s
\end{align}
\textcolor{red}{Assuming the matrix norm of the Jacobian, $\mathcal{J}(\sigma) \, W_{hh}$, is upper bounded by $\mu$, we have the inequality above which states that the upper bound of the cost derivative norm decays by a factor of $\mu$ based on how high $s$ is.}
In other words, increasing the size of the context ($s$), any perturbations of the input far back in context do not propagate to the loss gradient.

\textbf{With Traveling Waves}

With the Traveling Waves in Equation \ref{eqn:PhiTheoretical}, we have $h(s) = \epsilon + \sum_{\mu=1}^s \sum_{i}^d (x(\mu))^i \, \psi_{(\mu-1)d+i}$. Here $\epsilon$ is any activity in addition to the wave activity.
We can similarly derive the gradient of the cost function with respect to the input at timestep 1.
\begin{align}
    \pdv{C}{x(1)} &= \pdv{C}{h(s+1)} \, \mathcal{J}(\sigma) \, W_{hh} \left( \sum_{i}^d \psi_{i} +  \pdv{\epsilon}{x(1)} \right) 
\end{align}
Similar to the previous analysis, we can upper bound the norms
\begin{align}
    \Vert \pdv{C}{x(1)} \Vert &\leq \Vert \pdv{C}{h(s+1)} \Vert \, \left( \mu d +  \mu \Vert \pdv{\epsilon}{x(1)} \Vert \right) \\ 
\end{align}
It can be noted here that compared to the analysis without TWM, there is no power on the $\mu d$ term to compute the gradient. There can be higher powers of $\mu$ in $\mu \Vert \pdv{\epsilon}{x(1)} \Vert$ terms, however, these terms contribute only additively to the upper bound. As a result, the gradient upper bound need not vanish for the traveling waves. 

In other words, since each of the RNN inputs $x(t)$ is stored explicitly spatially in the hidden state as the activity of propagating waves, how far back the input is, does not effect gradient computation. \textcolor{red}{This holds as long as the gradient is propagated within the limits of the wave substrate. The experimental results in Figure \ref{fig:VMGradientPropagation} show the gradient propagation issue indicated by the gradient norm improving during training as the traveling waves are learned in the RNN weights}.

\section{Discussion}

In this work, we introduced a Traveling wave based Working Memory (TWM), that stores variables required for computation in neural waves of activity instead of fixed register-like locations.
We defined two cases of the theory where the variables are updated linearly using linear boundary conditions (LBC) and non-linearly using self-attention based boundary conditions (SBC). We theoretically connected LBC to Recurrent Neural Networks shedding light into the potential information storage mechanisms in these networks.
We also connected SBC to transformer architectures, providing an alternate justification for the computational advantages of transformers over standard recurrent architectures.

The investigation of empirical RNNs trained on history-dependent dynamical systems revealed that they employ hidden traveling waves for the storage of external information.
Building on the evidence, we defined a linear basis for the trained RNN parameters that revealed latent waves actively involved in working memory storage.
Further, we viewed the learned parameters of an RNN in a human-interpretable manner, enabling reasoning and understanding RNN behavior as propagating wave activity.
Using the tools from the theory, we \textit{fully} deconstructed both the hidden state behavior \textit{and} the learned parameters of empirically trained RNNs.
We further showed that traveling waves have improved learning behavior by alleviating the diminishing/exploding gradient issue.
\textcolor{red}{A notable recent work \cite{Keller2023TravelingWE} showed empirical success by explicitly introducing waves into RNN computation. In their experiments they showed how the RNN with the explicitly added wave dynamics train efficiently, and generalize better compared to other RNN variants.
Our work adds theoretical support and adds empirical evidence that wave computation \textit{emerges} naturally in simple RNN networks without any modifications.}
Through both theoretical and empirical evidence, we add credibility to the claim that traveling waves play a role in information storage in biological \textit{and} artificial intelligence.
%
% Additional investigation and experiments will reveal the extend to which different wave behaviors influence wave computation.

% As evidence increases on the utility of traveling waves in neurobiology, we add additional 

% The Episodic Memory Theory and variable memories are versatile enough to be broadly applicable in various scenarios, offering valuable insights for researchers designing new algorithms. One practical application of this analysis can be in the development of continual learning algorithms, which can restrict gradients to pre-existing variable memory spaces to minimize catastrophic forgetting of prior tasks. Additionally, in task composition—where a new task is a combination of two existing tasks—the linear spaces of each task can be linearly combined to efficiently solve the composite problem. Another possible application is transfer learning, where task knowledge is shared between networks. The Episodic Memory Theory suggests that variable memories and their interactions are the essential components for knowledge transfer, allowing the remaining network dynamics to be disregarded or easily relearned, streamlining the transfer process.

\textbf{Limitations}: With these results, it is also important to recognize inherent limitations to the traveling wave approach. One of the limitations is that the analysis using linear basis transformations we presented is primarily restricted to linear and binary HDS.  Although an accurate representation of the qualitative behavior within \textit{small neighborhoods} of fixed points can be found for non-linear dynamical systems \cite{Sussillo2013OpeningTB}, the RNNs have to be confined to these linear regions for the analysis to be applicable. 
It is still an interesting behavior that models consistently converge to the linear regime, at least for the restricted tasks we consider in the paper.
\textcolor{red}{Generalizing the basis transformation for the case of non-linear tasks will enable the understanding of wave computation to be used directly to improve and understand neural computation.}
The second limitation of the approach is that the external information is stored as a \textit{linear} basis in the hidden state. Our results indicates that the role of non-linearity in encoding external information may be minimal for these elementary HDS. However, we have observed that when the number of dimensions of the linear operator $W_{hh}$ is not substantially large compared to the task's dimensionality requirements (bottleneck superposition) or when the regularization penalty is high, the RNN can effectively resort to non-linear encoding mechanisms to store external information (Appendix \ref{appendix:ERVB:FurtherExamples}).
Overcoming the limitations of non-linearity will be an interesting direction to pursue in future research, and will bring the concept of traveling waves in addressing the challenges posed by the quest to understand neural computation.

%% file: appendix.tex
\section{Theoretical Models of Variable Binding} \label{appendix:section:TheoreticalModelsOfVariableBinding}

In the appendix, we use the Dirac and Einstein summation conventions for brevity to derive all the theoretical expositions in the main paper. 

\subsection{Mathematical Preliminaries} \label{appendix:TMVB:MathematicalPreliminaries}
The core concept of the episodic memory theory is basis change, the appropriate setting of the stored memories. Current notations lack the ability to adequately capture the nuances of basis change.
Hence, we introduce abstract algebra notations typically used in theoretical physics literature to formally explain the variable binding mechanisms. We treat a \textit{vector} as an abstract mathematical object invariant to basis changes. 
Vectors have \textit{vector components} that are associated with the respective basis under consideration. We use Dirac notations to represent vector $v$ as - $\ket{v} = \sum_i v^i \ket{e_i}$.
Here, the linearly independent collection of vectors $\ket{e_i}$ is the \textit{basis} with respect to which the vector $\ket{v}$ has component $v^i \in \mathbb{R}$. Linear algebra states that a collection of basis vectors $\ket{e_i}$ has an associated collection of \textit{basis covectors} $\bra{e^i}$ defined such that $\bra{e^i} \ket{e_j} = \delta_{i j}$,
where $\delta_{i j}$ is the Kronecker delta. This allows us to reformulate the vector components in terms of the vector itself as $\ket{v} = \sum_{i} \bra{e^i} \ket{v} \ket{e_i}$.
We use the Einstein summation convention to omit the summation symbols wherever the summation is clear. Therefore, vector $\ket{v}$ written in basis $\ket{e_i}$ is
\begin{dmath}
    \ket{v} = v^i \ket{e_i} = \bra{e^i} \ket{v} \ket{e_i} .
\end{dmath}
The set of all possible vectors $\ket{v}$ is a \textit{vector space} spanned by the basis vectors $\ket{e_i}$. A \textit{subspace} of this space is a vector space that is spanned by a subset of basis vectors $\{ \ket{e^\prime_j}:  \ket{e^\prime_j} \subseteq \{ \ket{e_i} \} \}$. 

\subsection{Traveling Wave Model (TWM)} 
\label{Appendix:TWMDefinition}
The traveling wave model is the discretization of the wave equation traveling in a neural lattice with only interactions at the boundaries.
This is a generalization of the model considered in the WaveRNN paper \cite{Keller2023TravelingWE}.
We consider independent $d$ dimensional waves traveling in a continuous medium (represented by $v_j$).
\begin{equation}
    \pdv{h(v_j, t)}{t} = \nu_j \pdv{h(v_j,t)}{x}
\end{equation}
The index $j \in \{ 1, 2, \hdots, d\}$. There is no interaction between the $d$ different waves traveling in the substrate except at the boundary location 1 (the beginnning of the wave):
\begin{equation}
    h_i(0, t) = f_i(h(0, t), h(1, t), \hdots, h(s, t))
\end{equation}
After Euler discretization of the system with step size $1$, 
\begin{equation}
    h(v_j, t+1) - h(v_j, t)  = \nu_j \left( h(v_j+1, t) - h(v_j, t) \right)
\end{equation}
For uniform wave propagation speeds $\nu_j = 1, \forall i$,
\begin{equation}
    h(v_j, t+1) - h(v_j, t)  = \left( h_i(v_j+1, t) - h_i(v_j, t) \right)
\end{equation}
Finally after simplifications, we obtain the following equations defining how the wave propagates in the neural substrate
\begin{dmath}
    \begin{dcases}
        h(v_j, t+1) = h(v_j+1, t) & \text{ when } 1 < k \leq s \\
        h(0, t+1) = f(h(0, t), h(1, t), \hdots, h(s, t)) & 
    \end{dcases}
\end{dmath}
Mapping this back to the notation that is used in the rest of the paper,
\begin{dmath}
    \begin{dcases}
        h_{i, j}(t+1) = h_{i+1, j}(t) & \text{ when } 1 < k \leq s \\
        h_{s, j}(t+1) = f_j(h_{i j}(t)) & \forall j \{1, 2, \hdots d\} \text{ and } i \in \{1, 2, \hdots s\} 
    \end{dcases}
\end{dmath}

The RNN dynamics presented in Equation \ref{eqn:LinearizedTWM} represented in the new notation is reformulated as: 
\begin{dmath} \label{eq:elmanRNN}
    \ket{h(t)} = \sigma_f \left( \left( \xi^i_{\mu} \, \Phi^\mu_\nu \, (\xi^\dag)^\nu_j \; \ket{e_i} \bra{e^j} \right) \ket{h(t-1)} \right) = \sigma_f \left( W_{hh} \, \vec{h}(t-1) \right) .
\end{dmath}
The greek indices iterate over memory space dimensions $\{ 1, 2, \hdots, N_h \}$, alpha numeric indices iterate over feature dimension indices $\{ 1, 2, \hdots, N_f \}$.
Typically, we use the standard basis in our simulations. For the rest of the paper, the standard basis will be represented by the collection of vectors $\ket{e_i}$ and the covectors $\bra{e^i}$. The hidden state at time $t$ in the standard basis is denoted as $\ket{h(t)} = \bra{e^j} \ket{h(t)} \ket{e_i}$.
$\bra{e^j} \ket{h(t)}$ are the \textit{vector components} of $\ket{h(t)}$ we obtain from simulations.

\subsection{Example: Generalized Fibonacci Series. } 
\label{apppendix:TMVB:fibonacci}
We consider a generalization of the Fibonacci sequence where each element $F_n \in \mathbb{R}^d$ is a vector defined recursively as,
\begin{equation}
    F_n = \begin{cases}
        u^1 & n=1 \\
        u^2 & n=2 \\
        \vdots & \\
        u^s & n=s \\
        \sum_{t=n-s}^{n-1} F_t & n > s
    \end{cases}
\end{equation}
For any $u^i \in \mathbb{R}^d$. In order to store this \textit{process} of generating sequences of $F_n$, the vectors $u_1, u_2, \hdots u_s$ needs to be stored as variables and recursively added to produce new outputs. 
In our framework, this can be accomplished by initializing the hidden state such that $\ket{h(s)} = \sum_i \sum_{\psi_\mu \in \{\Psi_i \}} u^\mu_i \, \ket{\psi_\mu}$, that is each $u^\mu$ is stored as activity of distinct subspaces of the hidden state.
To encode the Fibonacci process in $\Phi$, we propose the following form for the inter-memory interactions.
\begin{dmath}
    \Phi = \sum_{\mu=1}^{(s-1) d} \ket{\psi_{\mu}} \bra{\psi^{\mu+d}} + \underbrace{\left( \sum_{\mu=1}^{d} \ket{\psi_{(s-1)d + \mu}} \left(\sum_{\nu=0}^{s-1} \bra{\psi^{\nu d + \mu}} \right) \right)}_{\sum_{t=n-s}^{n-1} F_t} \, .
\end{dmath}
This form of $\Phi$ has two parts. The first part implements a variable shift operation.
The second part implements the summation function of Fibonacci. 
Since the hidden state is initialized with all the starting variables $u^\mu$, application of the $\Phi$ operator repeatedly, produces the next element in the sequence. As of now, the hidden state contains all the elements in the sequence. 
The abstract algebra notation allows proposing $W_r$ which will extract only the required output. Formally,
\begin{equation}
    W_{r} = \Psi^*_s = \sum_{\mu=(s-1)d + 1}^{s d} \ket{e_{\mu - (s-1)d}} \bra{\psi^{\mu}} \, .
\end{equation}
It is the projection operator which extracts the contents of the $N^{\text{th}}$ variable memory in the standard basis.
Note that the process works irrespective of the actual values of $u^{\mu}$. 
To summarize, we now have a memory model encoding a generalizable process of fibonacci sequence generation.

\subsection{Example: Repeat Copy} \label{appendix:TMVB:example}

Repeat Copy is a task typically used to evaluate the memory storage characteristics of RNNs since the task has a deterministic evolution represented by a simple algorithm that stores all input vectors in memory for later retrieval.
Although elementary, repeat copy provides a simple framework to work out the variable binding circuit we theorized in action.
For the repeat copy task, the linear operators of the RNN has the following equations.
\begin{empheq}{equation}
    \begin{cases}
        \Phi = \sum_{\mu=1}^{(s-1) d} \ket{\psi_{\mu}} \bra{\psi^{\mu+d}} + \sum_{\mu=(s-1)d + 1}^{s d} \ket{\psi_\mu} \bra{\psi^{\mu - (s-1) d}} \\
        W_{uh} = \Psi_{s} \\
        W_r = \Psi^*_{s}
    \end{cases}
\end{empheq}
This $\phi$ can be imagined as copying the contents of the subspaces in a cyclic fashion. That is, the content of the $i^{th}$ subspace goes to $(i-1)^{\text{th}}$ subspace with the first subspace being copied to the $N^{\text{th}}$ subspace.
%
%For simplicity, lets assume that the stored memories are the standard basis vectors, that is $\ket{\psi_{\mu}} = \delta_{i \mu} \ket{e_{i}}$.
The dynamical evolution of the RNN is represented at the time step $1$ as,
\begin{dmath}
    \ket{h(1)} = \ket{\psi_{(s-1)d + j}} \bra{e^j} u^i(1) \ket{e_i}
\end{dmath}
\begin{dmath}
    \ket{h(1)} = u^i(1) \ket{\psi_{(s-1)d + j}} \bra{e^j} \ket{e_i}
\end{dmath}
\begin{dmath}
    \ket{h(1)} = u^i(1) \ket{\psi_{(s-1)d + j}} \delta_{i j}
\end{dmath}
Kronecker delta index cancellation
\begin{dmath}
    \ket{h(1)} = u^i(1) \ket{\psi_{(s-1)d + i}}
\end{dmath}
At time step $2$,
\begin{dmath}
    \ket{h(2)} = u^i(1) \, \Phi \, \ket{\psi_{(s-1)d + i}} + u^i(2) \, \ket{\psi_{(s-1)d + i}}
\end{dmath}
Expanding $\Phi$
\begin{dmath}    
    \ket{h(2)} = u^i(1) \, \left( \sum_{\mu=1}^{(s-1) d} \ket{\psi_{\mu}} \bra{\psi^{\mu+d}} + \sum_{\mu=(s-1)d + 1}^{s d} \ket{\psi_\mu} \bra{\psi^{\mu - (s-1) d}} \right) \, \ket{\psi_{(s-1)d + i}} + u^i(2) \, \ket{\psi_{(s-1)d + i}}
\end{dmath}
\begin{dmath}    
    \ket{h(2)} = u^i(1) \, \ket{\psi_{(s-2)d + i}} + u^i(2) \, \ket{\psi_{(s-1)d + i}}
\end{dmath}
At the final step of the input phase when $t=s$, $\ket{h(s)}$ is defined as:
\begin{dmath}    
    \ket{h(s)} = \sum_{\mu=1}^s u^i(\mu) \, \ket{\psi_{(\mu-1)d + i}}
\end{dmath}
For $t$ timesteps after $s$, the general equation for $\ket{h(s+t)}$ is:
\begin{dmath}    
    \ket{h(s+t)} = \sum_{\mu=1}^s u^i(\mu) \, \ket{\psi_{\left[ \left(\left(\mu-t-1 \mod s\right) + 1 \right)d + i \right]}}
\end{dmath}
From this equation for the hidden state vector, it can be easily seen that the $\mu^{\text{th}}$ variable is stored in the $\left[(\mu-t-1 \mod s) + 1\right]^{\text{th}}$ subspace at time step $t$. The readout weights $W_r = \Psi_s^*$ reads out the contents of the $s^{\text{th}}$ subspace.

\subsection{Application to General RNNs} \label{appendix:TMVB:generalRNNs}

The linear RNNs we discussed are powerful in terms of the content of variables that can be stored and reliably retrieved. The variable contents, $u^i$, can be any real number and this information can be reliably retrieved in the end using the appropriate readout weights.
However, learning such a system is difficult using gradient descent procedures. To see this, setting the components of $\Phi$ to anything other than unity might result in dynamics that is eventually converging or diverging resulting in a loss of information in these variables.
Additionally, linear systems are not used in the practical design of RNNs. The main difference is now the presence of the nonlinearity. In this case, our theory can still be used. 
To illustrate this, consider a general RNN evolving according to $h(t+1) = g(W_{hh} h(t) + b)$ where $b$ is a bias term. Suppose $h(t) = h^*$ is a fixed point of the system. We can then linearize the system around the fixed point to obtain the linearized dynamics in a small region around the fixed point. 
\begin{dmath}
    h(t+1) - h^* = \mathcal{J}(g)|_{h^*} \, W_{hh} \, (h(t+1) - h^*) + O((h(t+1) - h^*)^2)
\end{dmath}
where $\mathcal{J}$ is the jacobian of the activation function $g$. If the RNN had an additional input, this can also be incorporated into the linearized system by treating the external input as a control variable
\begin{dmath}
    h(t+1) - h^* = \mathcal{J}(g)|_{h^*} \, W_{hh} \, (h(t) - h^*) + \mathcal{J}(g)|_{h^*} \, W_{uh} u(t)
\end{dmath}
Substituting $h(t) - h^* = h^{\prime}(t)$
\begin{dmath}
    h^{\prime}(t+1) = \mathcal{J}(g)|_{h^*} \, W_{hh} \, h^{\prime}(t) + \mathcal{J}(g)|_{h^*} \, W_{uh} u(t)
\end{dmath}
which is exactly the linear system which we studied where instead of $W_{hh} = \Xi \Phi \Xi^\dag$, we have $J(g)|_{h^*} W_{hh} = \Xi \Phi \Xi^\dag$. 
%
% With this result, we will analyse Elman RNN models that have the general update equations $h(t+1) = \tanh(W_{hh} h(t) + W_{uh} u(t) + b)$.

\subsubsection{Fixed Point Finding Algorithm}

In order to perform the linearization analysis, we have to first identify the fixed points of the trained RNN. We utilized the procedure introduced in \cite{Sussillo2013OpeningTB} to obtain these fixed points. Specfically, we find fixed point $h^*$ by finding the solution to the following optimization problem:
\begin{equation}
    h^* = \arg \min_{x} \lVert g(W_{hh} \, x + b) - x \rVert
\end{equation}
Intuitively, the optimization function $\lVert g(W_{hh} \, x + b) - x \rVert$ reaches the minimum at the fixed point $h^*$.

\subsection{Error Analysis of the Variable Memory Approximation Algorithm} \label{appendix:algorithmFailureModes}

Our empirical results revealed that there are certain cases of tasks where the algorithm fails to retrieve the correct basis transformation. In this section, we will investigate why this dissociation from theory happens.
To this end, we want to formalize and compare what the hidden state is according to the power iteration ($h(t)$) and the variable memories ($\ket{h(t)}$).
\begin{equation}
    \ket{h(0)} = 0 \qquad h(0) = 0
\end{equation}
\begin{equation}
    \ket{h(1)} = u^i(1) \ket{\psi_i} \qquad h(1) = W^\dag_r u(1)
\end{equation}
\begin{equation}
    \ket{h(1)} = u^i(1) \ket{\psi_{(d+i)}} + (u^i(2) + \Phi (0, \hdots, u(1))^i) \ket{\psi_{i}} \qquad h(1) = W_{hh} W^\dag_r u(1) + W^\dag_r u(2)
\end{equation}
The error in the basis definition is given by
\begin{equation}
    \Tilde{\Psi}_2 - \Psi_2 = \Phi(0, \hdots u(1))^i \ket{\psi_i} \bra{e^j}
\end{equation}
\begin{equation}
    \Tilde{\Psi}_3 - \Psi_3 = \Phi(0, \hdots u(1))^i \ket{\psi_{(d+i)}} \bra{e^j} + \Phi(0, \hdots u(2), u(1))^i \ket{\psi_{(i)}} \bra{e^j} 
\end{equation}
For the power iteration to succeed in giving the correct variable memories, the effect of $\Phi$ acting on the variable memories needs to be negligible. In the case of repeat copy, this effect is zero as the operator $\Phi$ does not utilize any of the variables until the end of the input phase.
For some of the compose copy tasks, we showed in the paper, this effect is negligible.
% %
Another way to think about this issue is that the $\Phi$ keeps on acting on the variable memories during the input phase and produces outputs even though the variables are not filled in fully yet.
% %
This behavior \textit{pollutes} the definition of variable memories using power iteration.
% %
If $\Phi$ is sufficiently representative, for instance, the operator associated with the Fibonacci generation, then after the first input is passed, the 2nd variable memory will be the sum of the 1st and 2nd variable memories. Future development to the algorithm to general tasks needs to take this figure out ways to get over this error.

\begin{figure*}[ht]
         \centering
		\includegraphics[width=0.73\textwidth]{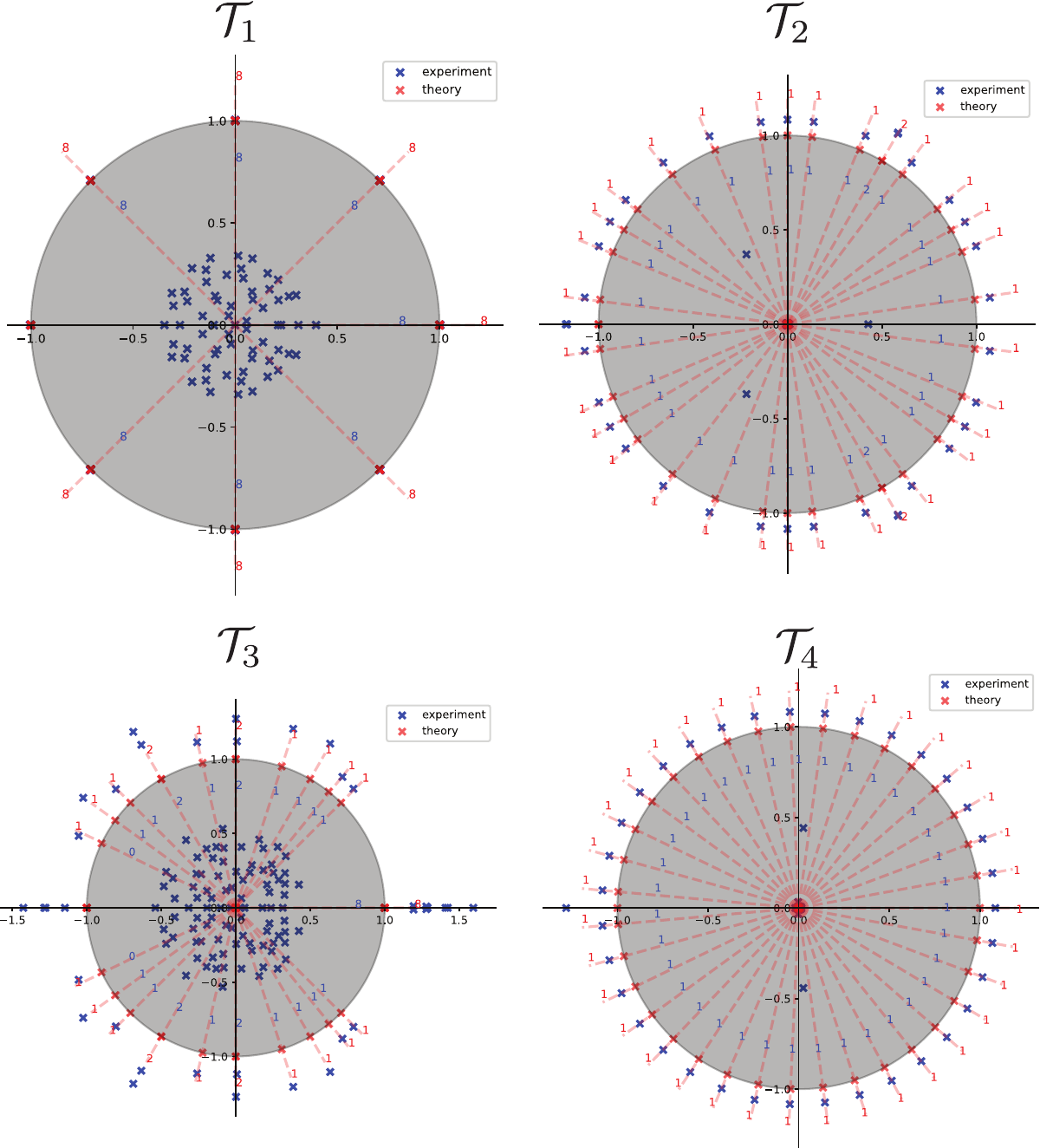}
\caption{ \textcolor{red}{\textbf{Diverse range of high-dimensional dynamical behavior around the fixed point for the variable binding tasks}: The figures show the distribution of the eigenspectrum of the Jacobian around origin for the Elman RNN in the complex plane. For each of the representative tasks, the spectral analysis reveals a very high dimensional dynamical behavior with a complex spectral distribution. The dynamical behavior is non-trivial to interpret by the analysis of the Jacobian spectrum in the class of variable binding tasks alone. Despite the diverse behaviors in the eigenspectrum, the complex argument of the eigenvalues are exactly predicted from the theory. The number of eigenvalues along each direction is counted and annotated on the lines.} }
\label{fig:appendix:diverseSpectrum}
\end{figure*}

\begin{figure*}[ht]
         \centering
		\includegraphics[width=0.73\textwidth]{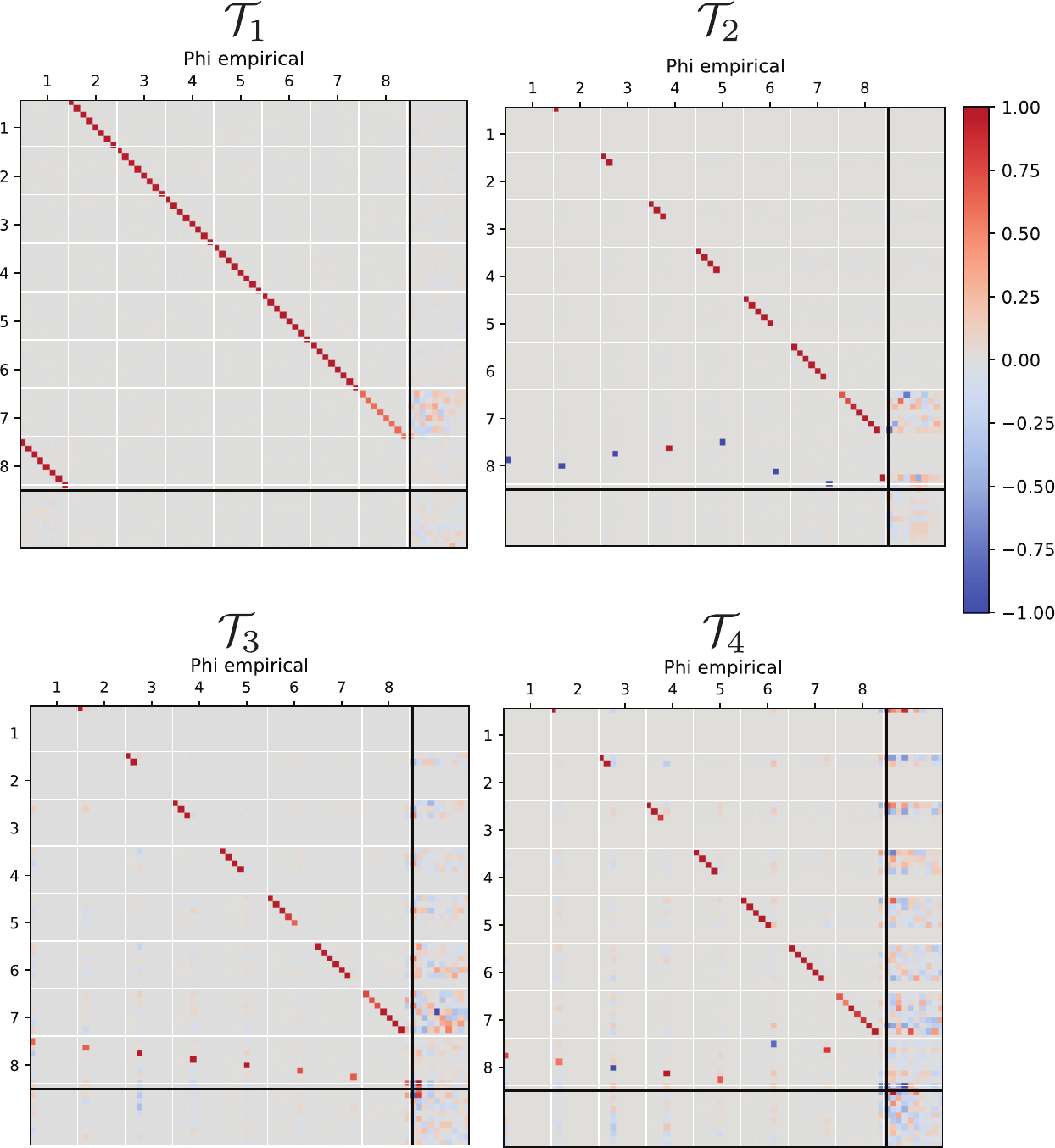}
\caption{ \textcolor{red}{\textbf{Transformed Learned RNN Parameters using the TWM theory}: The plots show 4 same RNNs learned on the 4 tasks in the paper. For all these tasks, the RNN consistently converged to the mechanisms predicted by the TWM theory.} }
\label{fig:appendix:connectivityTransformed}
\end{figure*}

\section{Experiments} \label{appendix:ERVB}

\subsection{Data} \label{appendix:ERVB:Data}

We train RNNs on the variable binding tasks described in the main paper with the following restrictions - the domain of $u$ at each timestep is binary $\in \{-1, 1\}$ and the function $f$ is a linear function of its inputs. We collect various trajectories of the system evolving accoding to $f$ by first sampling uniformly randomly, the input vectors.
The system is then allowed to evolve with the recurrent function $f$ over the time horizon defined by the training algorithm.

%%%%%%%%%%%%%%%%%%%%%%%%%%%%%%%%%%%%%%%%%%%%%%%%%%%%%%%%
%%%%%%%%%%%%%%%%%%%%%%%%%%%%%%%%%%%%%%%%%%%%%%%%%%%%%%%%
\subsection{Training Details} \label{appendix:ERVB:TrainingDetails}

\textbf{Architecture} \hspace{0.25cm} 
We used single layer Elman-style RNNs for all the variable binding tasks. Given an input sequence $(u(1), u(2), ..., u(T))$ with each $u(t)\in \mathbb{R}^d$, an Elman RNN produces the output sequence $y = (y(1),...,y(T))$ with $y(t) \in \mathbb{R}^{N_{out}}$ following the equations
\begin{equation}
h(t+1) = \tanh(W_{hh} h(t) + W_{uh} u(t)) \quad,\quad y(t) = W_r h(t)
\end{equation}
Here $W_{uh} \in \mathbb{R}^{N_h \times N_{in}}$, $W_{hh} \in \mathbb{R}^{N_h \times N_h}$, and $W_r \in \mathbb{R}^{N_{out} \times N_h}$ are the input, hidden, and readout weights, respectively, and $N_h$ is the dimension of the hidden state space. 
%
% We set $N_h = 128$ for the Repeat Copy Task and $N_h = 256$ for the Compose Copy task. 

The initial hidden state $h(0)$ for each model was \textit{not} a trained parameter; instead, these vectors were simply generated for each model and fixed throughout training. We used the zero vector for all the models.

\vspace{0.25cm}
\noindent \textbf{Task Dimensions} \hspace{0.25cm} Our results presented in the main paper for the repeat copy $(\mathcal{T}_1)$ and compose copy ($\mathcal{T}_2$) used vectors of dimension $d = 8$ and sequences of $s = 8$ vectors to be input to the model.

\vspace{0.25cm}
\noindent \textbf{Training Parameters} \hspace{0.25cm}
We used PyTorch's implementation of these RNN models and trained them using Adam optimization with the MSE loss. We performed a hyperparameter search for the best parameters for each task --- see table \ref{table:training-params} for a list of our parameters for each task. 

% Table with all hyperparameters
\begin{table}[ht]
\begin{center}
\begin{tabular}{| c | c | c |}
\hline
 & Repeat Copy ($\mathcal{T}_1$) & Compose Copy ($\mathcal{T}_2$) \\ \hline
 Input \& output dimensions & 8 & 8 \\ \hline
 Input phase (\# of timesteps) & 8 & 8 \\ \hline
 training horizon & 100 & 100 \\ \hline
Hidden dimension $N_h$ & 128 & 128 \\ \hline
 \# of training iterations & 45000 & 45000 \\ \hline
 (Mini)batch size & 64 & 64  \\ \hline
 Learning rate & $10^{-3}$ & $10^{-3}$  \\ \hline
 % Learning rate decay factor & 0.1 & none \\
 applied at iteration & 36000 & \\ \hline
 Weight decay ($L^2$ regularization) & none & none \\ \hline
 Gradient clipping threshold & $1.0$ & 1.0 \\ \hline
\end{tabular}
\caption{Architecture, Task, \& Training Parameters}
\label{table:training-params}
\end{center}
\end{table}
\noindent \textbf{Curriculum Time Horizon} \hspace{0.25cm} 
When training a network, we adaptively adjusted the number of timesteps after the input phase during which the RNN's output was evaluated. We refer to this window as the \textit{training horizon} for the model. 

Specifically, during training we kept a rolling average of the model's \textit{loss by timestep} $L(t)$, i.e. the accuracy of the model's predictions on the $t$-th timestep after the input phase. This metric was computed from the network's loss on each batch, so tracking $L(t)$ required minimal extra computation. 

The network was initially trained at time horizon $H_0$ and we adapted this horizon on each training iteration based on the model's loss by timestep. Letting $H_n$ denote the time horizon used on training step $n$, the horizon was increased by a factor of $\gamma = 1.2$ (e.g. $H_{n+1} \gets \gamma H_n$) whenever the model's accuracy $L(t)$ for $t \le H_{\textrm{min}}$ decreased below a threshold $\epsilon = 3 \cdot 10^{-2}$. Similarly, the horizon was reduced by a factor of $\gamma$ is the model's loss was above the threshold ($H_{n+1} \gets H_n / \gamma$). We also restricted $H_n$ to be within a minimum training horizon $H_0$ and maximum horizon $H_{\textrm{max}}$. These where set to 10/100 for the repeat copy task and 10/100 for the compose copy task.

We found this algorithm did not affect the results presented in this paper, but it did improve training efficiency, allowing us to run the experiments for more seeds.

%%%%%%%%%%%%%%%%%%%%%%%%%%%%%%%%%%%%%%%%%%%%%%%%%%%%%%%%
%%%%%%%%%%%%%%%%%%%%%%%%%%%%%%%%%%%%%%%%%%%%%%%%%%%%%%%%
\subsection{Repeat Copy: Further Examples of Hidden Weights Decomposition} \label{appendix:ERVB:FurtherExamples}

This section includes additional examples of the hidden weights decomposition applied to networks trained on the repeat copy task. 

\begin{figure*}[ht]
         \centering
		\includegraphics[width=0.73\textwidth]{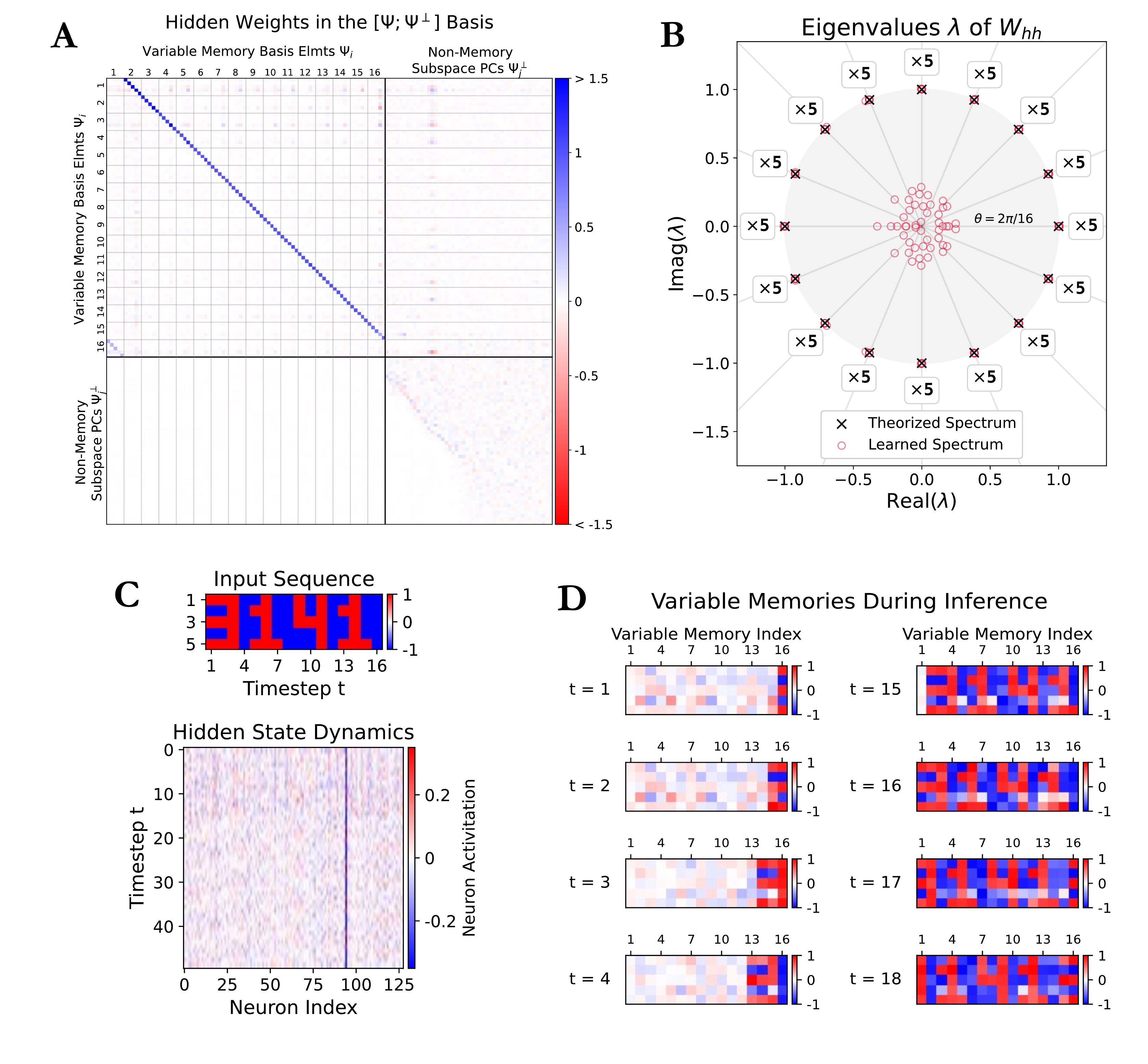}
\caption{ \textbf{Additional Experimental Results of Repeat Copy Task with 16 vectors, each of 5 dimensions}: \textbf{A}. $W_{hh}$ visualized in the variable memory basis reveals the variable memories and their interactions. \textbf{B}. After training, the eigenspectrum of $W_{hh}$ with a magnitude $\geq 1$ overlaps with the theoretical $\Phi$. The boxes show the number of eigenvalues in each direction.
\textbf{C}. During inference, "3141" is inserted into the network in the form of binary vectors. This input results in the hidden state evolving in the standard basis, as shown. How this hidden state evolution is related to the computations cannot be interpreted easily in this basis.
\textbf{D}. When projected on the variable memories, the hidden state evolution reveals the contents of the variables over time. Note that in order to make these visualization clear, we needed to normalize the activity along each variable memory to have standard deviation 1 when assigning colors to the pixels. The standard deviation of the memory subspaces varies due to variance in the strength of some variable memory interactions. These differences in interaction strengths does not impede the model's performance, however, likely due to the nonlinearity of the activation function. Unlike the linear model, interaction strengths well above 1 cannot cause hidden state space to expand indefinitely because the tanh nonlinearity restricts the network's state to $[-1,1]^{N_h}$. This property appears to allow the RNN to sustain stable periodic cycles for a range of interaction strengths above 1.}
\label{fig:RC-S16-d5-example2}
\end{figure*}

\begin{figure*}[ht]
         \centering
		\includegraphics[width=0.9\textwidth]{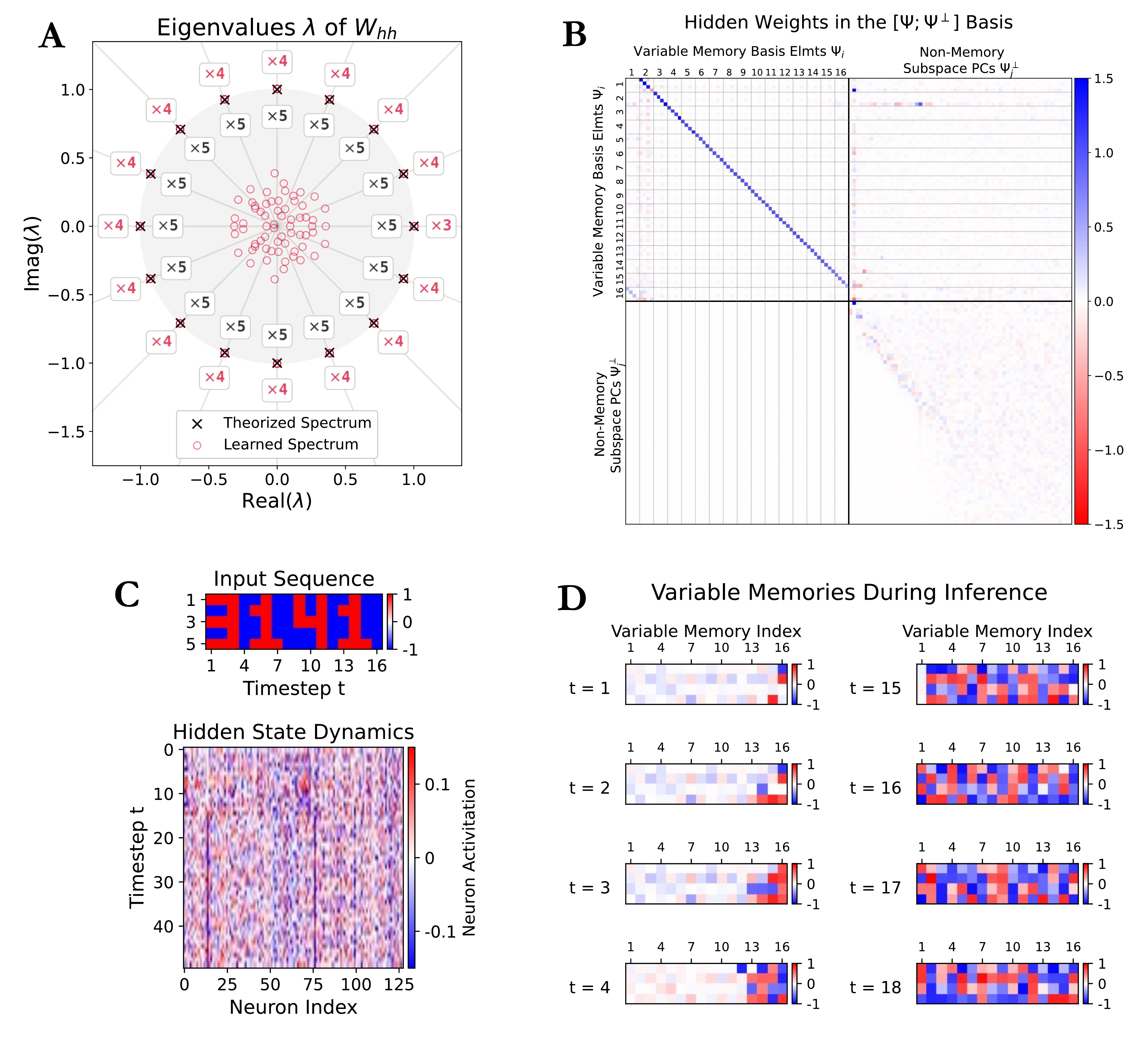}
\caption{ \textbf{Nonlinear Variable Memories Learned for the Repeat Copy Task with 16 vectors, each of 5 dimensions}. \textbf{A}. Eigenspectrum of $W_{hh}$ after training. The learned eigenvalues cluster into 16 groups equally spaced around the unit circle, but there are only 3-4 eigenvectors per cluster (indicated in red). Compare this to the theorized linear model, which has 5 eigenvalues per cluster (indicated in black). The task requires $16 \cdot = 5 = 80$ bits of information to be stored. Linearization about the origin predicts that the long-term behavior of the model is dictated by the eigenvectors with eigenvalue outside the unit circle because its activity along other dimensions will decay over time. The model has only $16 \cdot \mathbf{4} - 1 = 63$ eigenvectors with eigenvalue near the unit circle, so this results suggests the model has learned a non-linear encoding that compresses 80 bits of information into 63 dimensions. 
\textbf{B}: $W_{hh}$ visualized in the variable memory basis reveals the variable memories and their interactions. Here, the variable memories have only 4 dimensions because the network has learned only 63 eigenvectors with eigenvalue near the unit circle. The variable memory subspaces also have non-trivial interaction with a few of the the non-memory subspaces.
\textbf{C}. During inference, "3141" is inserted into the network in the form of binary vectors. This input results in the hidden state evolving in the standard basis, as shown. How this hidden state evolution is related to the computations cannot be interpreted easily in this basis.
\textbf{D}. When projected on the variable memories, the hidden state evolution is still not easily interpreted for this network, likely due to a nonlinear variable memories. As in the previous figure, we normalized the activity along each variable memory to have standard deviation 1 when assigning colors to the pixels.}
\label{fig:RC-S16-d5-nonlinear}
\end{figure*}

\begin{figure*}[ht]
         \centering
		\includegraphics[width=0.9\textwidth]{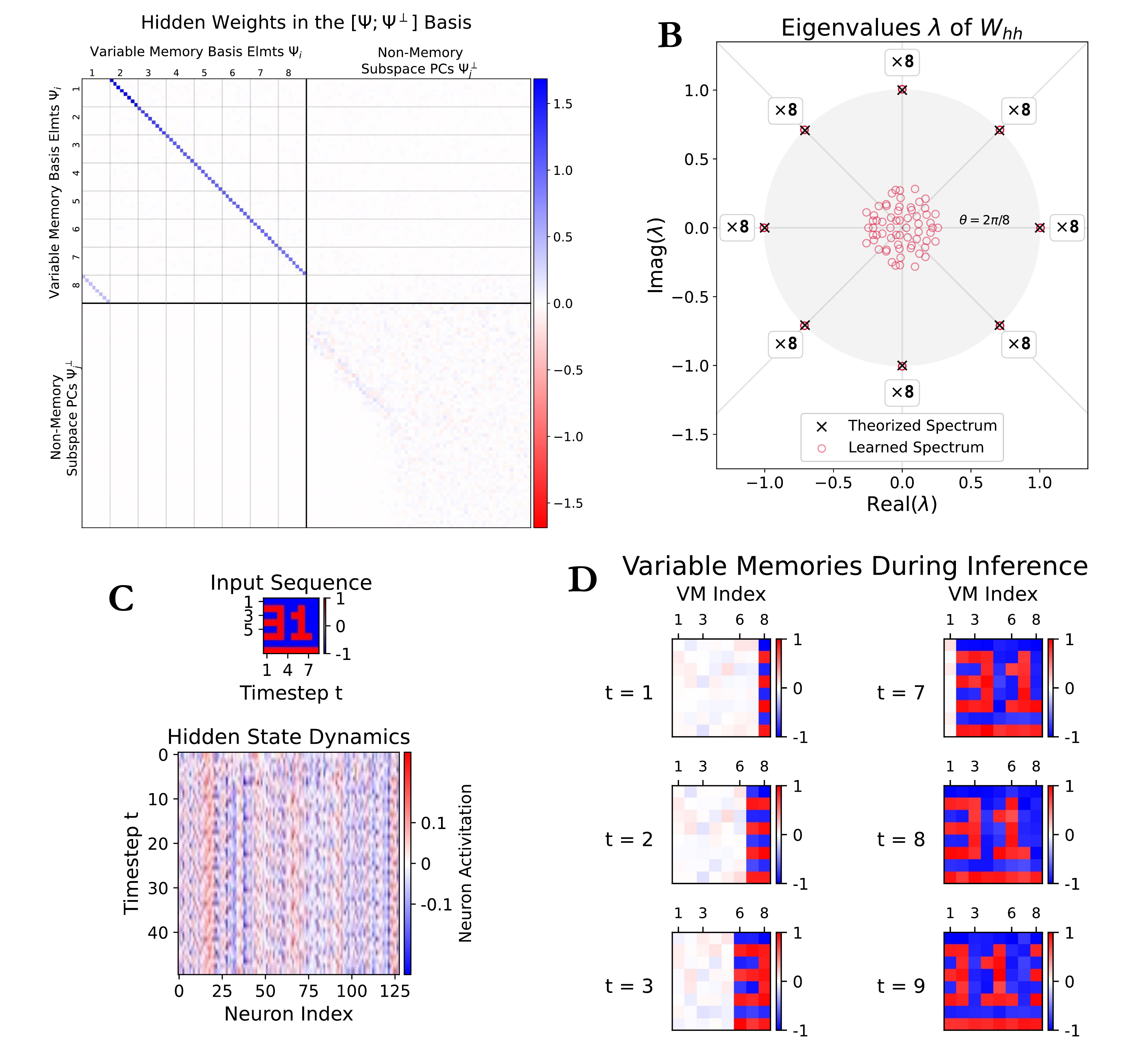}
\caption{ \textbf{Additional Experimental Results of Repeat Copy Task with 8 vectors, each of 8 dimensions}. This figure was included to show the decomposition applied to other values of $s$ and $d$ for the Repeat Copy task. \textbf{A}: $W_{hh}$ visualized in the variable memory basis reveals the variable memories and their interactions. \textbf{B}. After training, the eigenspectrum of $W_{hh}$ with a magnitude $\geq 1$ overlaps with the theoretical $\Phi$. The boxes show the number of eigenvalues in each direction.
\textbf{C}. During inference, "3141" is inserted into the network in the form of binary vectors. This input results in the hidden state evolving in the standard basis, as shown. How this hidden state evolution is related to the computations cannot be interpreted easily in this basis.
\textbf{D}. When projected on the variable memories, the hidden state evolution reveals the contents of the variables over time. As in the previous figures, we normalized the activity along each variable memory to have standard deviation 1 when assigning colors to the pixels.}
\label{fig:RC-S8-d8-example2}
\end{figure*}

\begin{figure*}[t]
         \centering
		\includegraphics[scale=0.82]{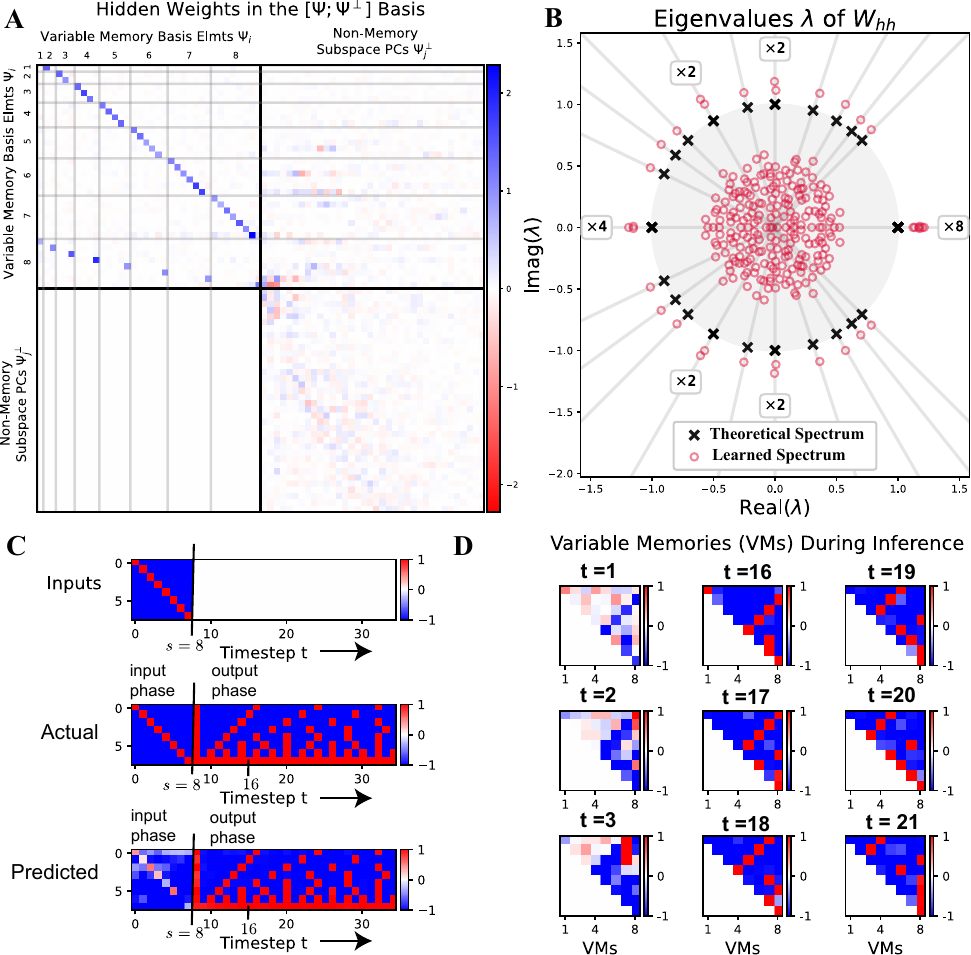}
\caption{ \textbf{Experimental Results of Compose Copy Task ($\mathcal{T}_3$) of 8 vectors, each of 8 dimensions}: \textbf{A}. $W_{hh}$ visualized in the variable memory basis reveals the variable memories and their interactions. It is observed that $W_{hh}$ encodes an optimized version of the theoretical mechanisms since there are dimensions in the variable memories irrelevant for future computations.
\textbf{B}. Compared to repeat copy, the eigenspectrum of the learned $W_{hh}$ is more complex, yet the theoretical $\Phi$ accurately predicts the angles and number of eigenvalues. The eigenvalues' magnitude greater than 1 (rather than close to 1 found in repeat copy) indicate that the non-linearity plays a role in controlling the diverging behavior of the spaces.
\textbf{C}. During the output phase, the past $s$ variables are composed to form future outputs. 
\textbf{D}. The hidden state evolution, when projected on the variable memories, reveals the contents of the variables over time. Unlike repeat copy, the input phase does not precisely match the model's theoretical predictions. Transient dynamics dominate the initial timesteps, clouding the underlying computations (t=1, 2, 3). Yet the long-term behavior (from t=16) of the output phase behavior is as the theoretical model predicts with the composed result stored in the $8^{\text{th}}$ variable memory at each time.  }
\label{fig:VB_ComposeCopy}
\end{figure*}

%%%%%%%%%%%%%%%%%%%%%%%%%%%%%%%%%%%%%%%%%%%%%%%%%%%%%%%%
%%%%%%%%%%%%%%%%%%%%%%%%%%%%%%%%%%%%%%%%%%%%%%%%%%%%%%%%
\clearpage
\subsection{Uniform vs. Gaussian Parameter Initialization}

\begin{figure*}[ht]
\centering
\begin{subfigure}
    \centering
    \includegraphics[scale=0.5]{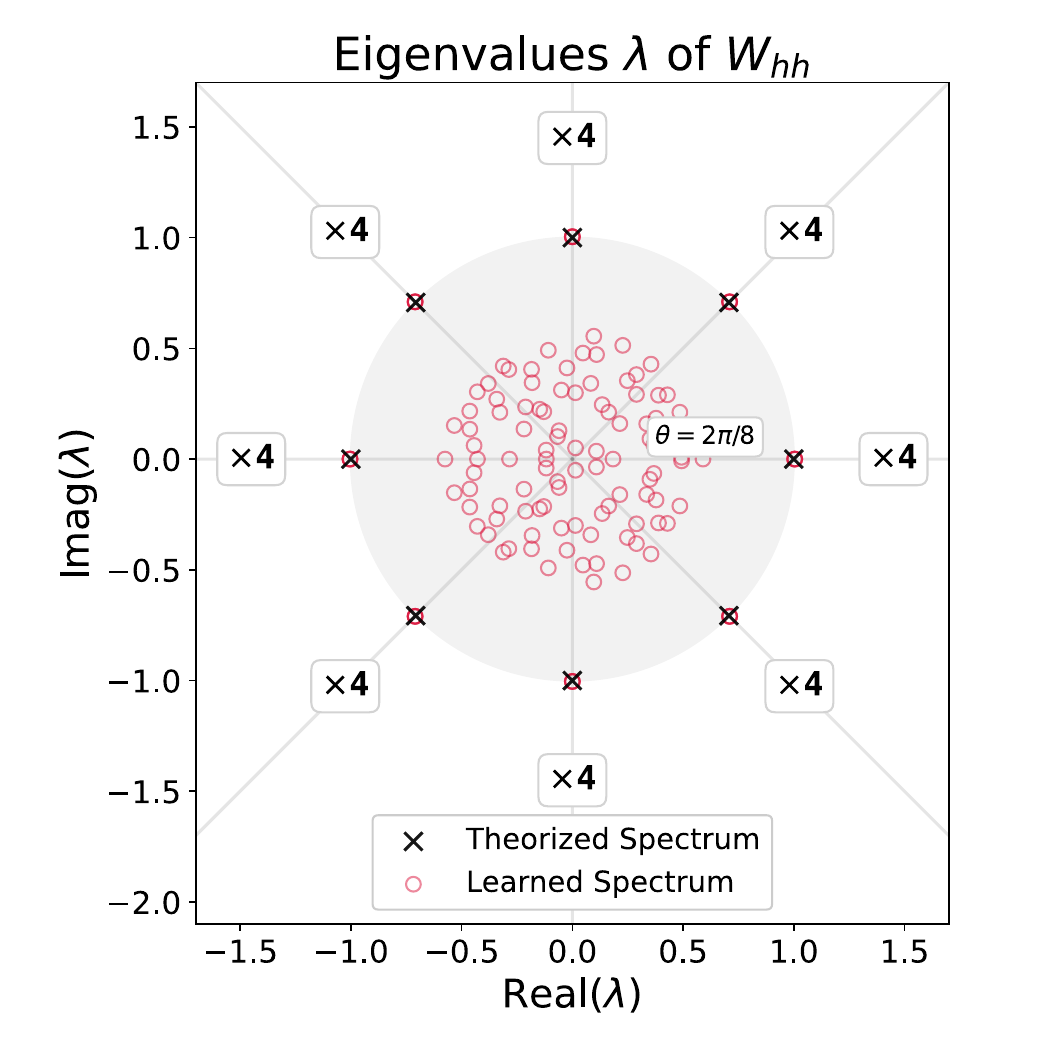}
    \caption{Uniform Initialization}
\end{subfigure}
\begin{subfigure}
    \centering
    \includegraphics[scale=0.5]{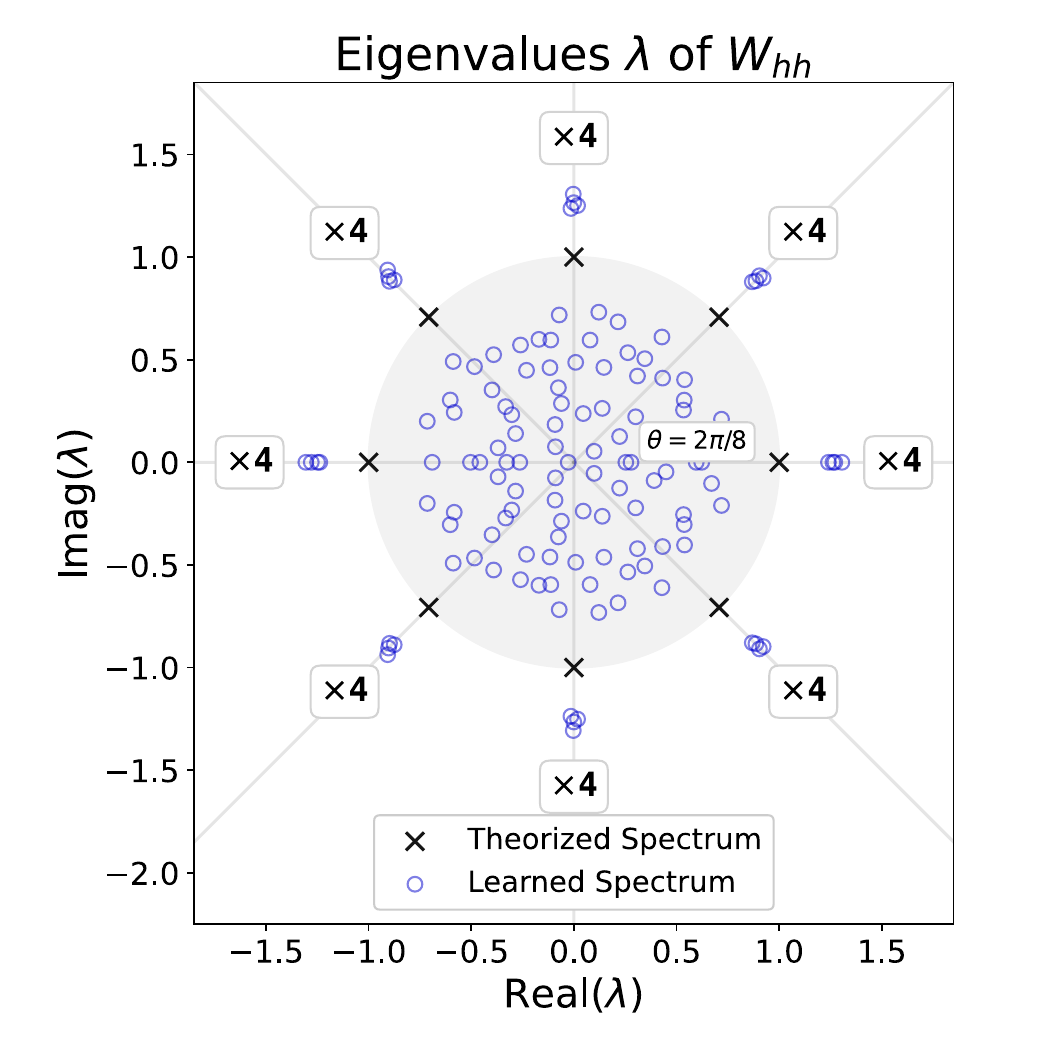}
    \caption{Gaussian Initialization}
\end{subfigure}
\caption{ \textbf{An effect of parameter initialization for the Repeat Copy Task} with $s = 8$ vectors, each of dimension $d = 4$. \textbf{A}: Spectrum (in red) of the learned hidden weights $W_{hh}$ for a network whose parameters where initialized from a uniform distribution over $[-k,k]$ with $k = 1/\sqrt{N_h}$. This network has 32 eigenvalues that are nearly on the unit circle. These eigenvalues are clustered into groups of 4, each group being an angle of $\theta = 2\pi/s$ apart from each other. These eigenvalues coincide with the eigenvalues of the linear model for solving the repeat copy task. \textbf{B}: Spectrum (in blue) of the learned hidden weights $W_{hh}$ for a network whose parameters where initialized from a Gaussian distribution with mean 0 and variance $1/N_h$. This network has 32 eigenvalues outside the unit circle, but they are a larger radii than the model initialized using the uniform distribution. These eigenvalues still cluster into eight groups of four, and the average complex argument of each group aligns with the complex arguments of the eigenvalues for the linear model.
}
\label{fig:RC_gaussian}
\end{figure*}

We also tested a different initialization scheme for the parameters $W_{uh}, W_{hh}$, and $W_{hy}$ of the RNNs to observe the effect(s) this would have on the structure of the learned weights. The results presented in the main paper and in earlier sections of the Supplemental Material used PyTorch's default initialization scheme: each weight is drawn \textit{uniformly} from $[-k,k]$ with $k = 1/\sqrt{N_h}$. Fig. \ref{fig:RC_gaussian} shows the resulting spectrum of a trained model when it's parameters where drawn from a Gaussian distribution with mean 0 and variance $1/N_h$. One can see that this model learned a spectrum similar to that presented in the main paper, but the largest eigenvalues are further away from the unit circle. This result was observed for most seeds for networks trained on  the repeat copy task with $s = 8$ vectors of dimension $d = 4$ and $d = 8$, though it doesn't hold for every seed. We also find that the networks whose spectrum has larger eigenvalues usually generalize longer in time than the networks with eigenvalues closer to the unit circle.

